\documentclass[journal,transmag]{IEEEtran}
\usepackage{times}
\usepackage{epsfig}
\usepackage{graphicx}
\usepackage{amsmath}
\usepackage{amssymb}
\usepackage{bbding}
\usepackage{subfigure}
\usepackage{xcolor}
\usepackage{flushend}
\usepackage{threeparttable}
\usepackage[pagebackref=true,breaklinks=true,letterpaper=true,colorlinks,bookmarks=false]{hyperref}
\newcommand{\para}[1]{\noindent{\textbf{#1}}}
\usepackage{bbding}
\usepackage{bbm}
\usepackage{bm}

\usepackage{expl3}
\ExplSyntaxOn
\newcommand\latinabbrev[1]{
	\peek_meaning:NTF . {
		#1\@}%
	{ \peek_catcode:NTF a {
			#1.\@ }%
		{#1.\@}}} 
\ExplSyntaxOff
\def\eg{\latinabbrev{e.g}}

\def\ie{\latinabbrev{i.e}}

\newcommand{\fig}[1]{Figure~#1}

\newtheorem{definition}{Definition}
\newtheorem{theorem}{Theorem}
\newtheorem{lemma}{Lemma}
\newtheorem{corollary}{Corollary}

\def\eg{\latinabbrev{e.g}}

\def\ie{\latinabbrev{i.e}}

\begin{document}

\title{Online Video Instance Segmentation via Robust Context Fusion}

\author{\IEEEauthorblockN{Xiang Li,
Jinglu Wang, Xiaohao Xu,
Bhiksha Raj, \IEEEmembership{Fellow,~IEEE}
Yan Lu, \IEEEmembership{Senior Member,~IEEE}
}
\thanks{
Part of this work was done when Xiang Li was an intern at Microsoft Research Asia.
Xiang Li is currently with Department of Electrical and Computer Engineering,
Carnegie Mellon University, Pittsburgh, PA 15213 USA. (email: xl6@andrew.cmu.edu)
}
\thanks{Jinglu Wang and Yan Lu are with Microsoft Research Asia, Beijing, China.  (email: jinglwa@microsoft; yanlu@microsoft.com)
}
\thanks{This work was done when Xiaohao Xu was an intern at Microsoft Research Asia.
Xiaohao Xu is currently with Department of Mechanical Engineering,
Huazhong University of Science and Technology, Wuhan, China. (email: xxh11102019@outlook.com)
}
\thanks{Bhiksha Raj is currently with School of Computer Science,
Carnegie Mellon University, Pittsburgh, PA 15213 USA. (email: bhiksha@cs.cmu.edu)
}
}

\markboth{}
{Shell \MakeLowercase{\textit{et al.}}: Bare Demo of IEEEtran.cls for IEEE Transactions on Magnetics Journals}

\IEEEtitleabstractindextext{%
\begin{abstract}

Video instance segmentation (VIS) aims at classifying, segmenting and tracking object instances in video sequences. Recent transformer-based neural networks have demonstrated their powerful capability of modeling spatio-temporal correlations for the VIS task. Relying on video- or clip-level input, they suffer from high latency and computational cost. We propose a robust context fusion network to tackle VIS in an online fashion, which predicts instance segmentation frame-by-frame with a few preceding frames. To acquire the precise and temporal-consistent prediction for each frame efficiently, the key idea is to fuse effective and compact context from reference frames into the target frame.  Considering the different effects of reference and target frames on the target prediction, we first summarize contextual features through importance-aware compression. A transformer encoder is adopted to fuse the compressed context. Then, we leverage an order-preserving instance embedding to convey the identity-aware information and correspond the identities to predicted instance masks. We demonstrate that our robust fusion network achieves the best performance among existing online VIS methods and is even better than previously published clip-level methods on the Youtube-VIS 2019 and 2021 benchmarks.

In addition, visual objects often have acoustic signatures that are naturally synchronized with them in audio-bearing video recordings. By leveraging the flexibility of our context fusion network on multi-modal data, we further investigate the influence of audios on the video-dense prediction task, which has never been discussed in existing works. We build up an Audio-Visual Instance Segmentation dataset, and demonstrate that acoustic signals in the wild scenarios could benefit the VIS task.

\end{abstract}

}

\maketitle

\IEEEdisplaynontitleabstractindextext

\IEEEpeerreviewmaketitle

\section{Introduction}
\label{sec:introduction}

The recently introduced video instance segmentation (VIS) problem receives increasing attention because of growing interest among researchers in the multimedia community. VIS aims at simultaneously classifying, segmenting and tracking objects in video sequences.  

With strong capabilities for modeling long-range data dependencies, some transformer-based methods \cite{vaswani2017attention,hwang2021video} achieve impressive results. However, they deal with the input at the video- or clip- level, thus incurring high latency. The cost of modeling full spatio-temporal correlations is prohibitive for practical applications.
In this work, we focus on \textit{online} VIS for streaming applications. The problem setting is as follows: given a small set of preceding \textit{reference} frames, the goal is to segment, classify and track object instances in each \textit{target} frame.

In addition to the challenges of segmentation and classification in the image domain, online VIS should also address the problem of finding correspondences and fusing contexts in adjacent frames. Yang et al. \cite{yang2019video} approach this problem by performing frame-based predictions independently and fusing the evidence across frames in a post-processing stage using sophisticated matching algorithms.
Matching as post-processing has a high cost, and the final result may suffer from flickering because of neglecting feature-level correlations across frames.
Several subsequent methods \cite{li2021spatial, athar2020stem,fu2020compfeat,qi2021occluded} fuse inter-frame features in the encoding stage. In particular, some methods \cite{fu2020compfeat,li2021spatial,qi2021occluded} crop out ROI features to fuse context within local regions, but cropped features are isolated from the global context.
To involve the global context, some methods \cite{fu2020compfeat,wang2021end,ke2021prototypical} generally concatenate reference and target feature maps together; such ``fusion'' procedures do not distinguish reference and target frames, and thus the specific importance of the target frame is ignored.
Moreover, as reference frames are usually similar to the target, spatio-temporal correlations among them could be highly redundant. Some reference features unrelated to the target could even mislead the target prediction. 

Besides the contextual information contained in reference video frames, information in other modalities, such as audio, can also serve as reference context to guide the VIS task. Visual objects often have acoustic signatures that are naturally synchronized with them in audio-bearing video recordings.
While audio-visual synchrony has been exploited in many contexts, most audio-visual representation learning methods handle signals in constrained environments, with clearly-evident audio-visual correspondences, such as music performance or lip reading. We deal with videos with audios in the wild where the correlation between audio and visual signals is difficult to establish. Previous methods \cite{zhao2018sound, tian2018audio, senocak2018learning, morrone2019face} have already demonstrated the correlation between audio and video modalities through sound localization and audio-visual separation, but there is no investigation on dense prediction done jointly with audio and visual modalities, potentially offering lost opportunities.
%


 

\begin{figure}
    \centering
    \includegraphics[width=\linewidth]{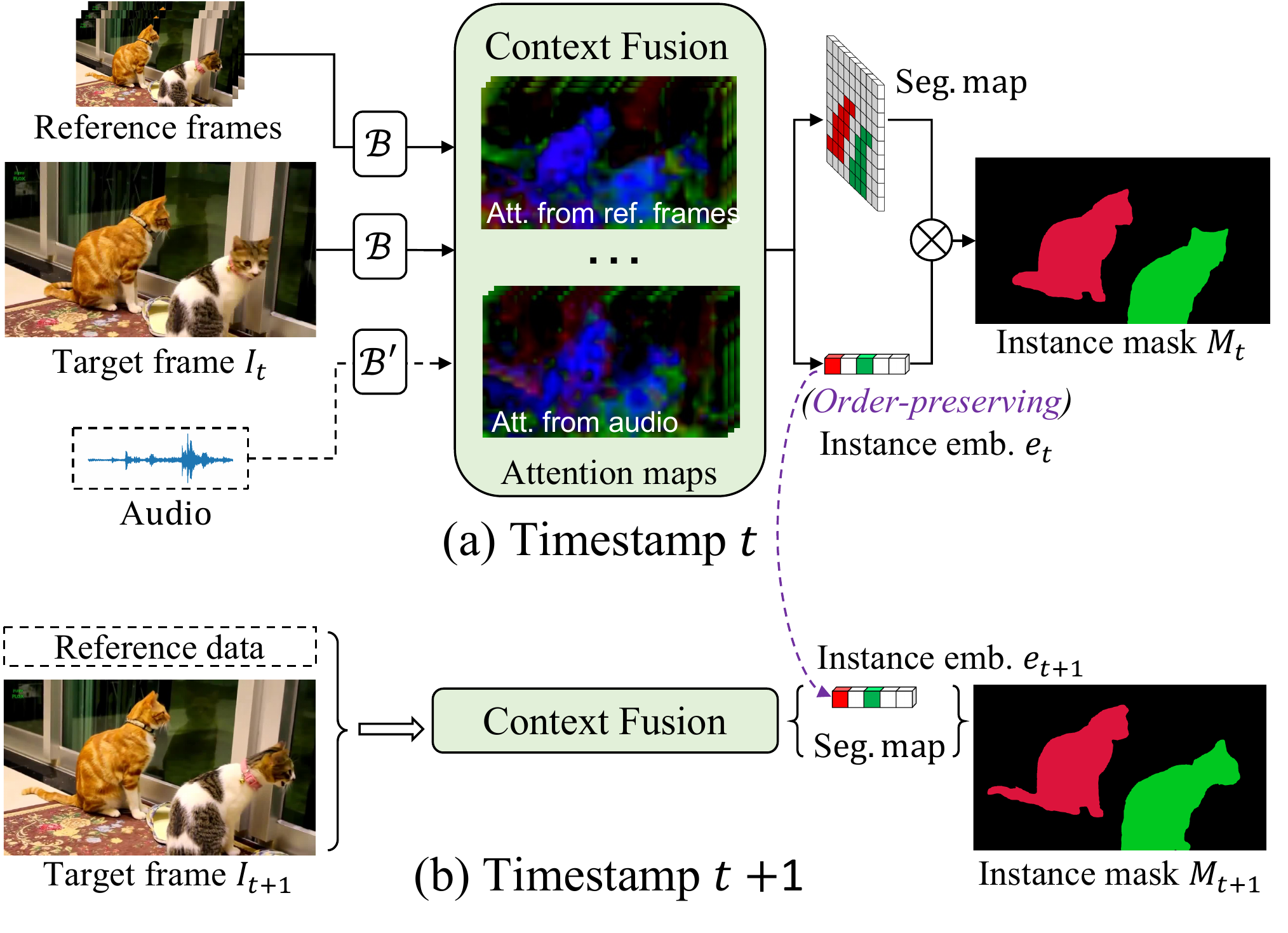}
    \caption{ 
    We propose a robust context fusion (RCF) network for the online VIS task. Considering the redundancy and noise in the reference visual or audio context, the RCF module extracts compact and representative contexts from features (extracted by backbones $\mathcal{B}$ and $\mathcal{B}'$), and then fuses them into the target feature with expressive attentions. The fused context is decoded into an instance code and segmentation maps. We directly compose the final instance masks by leveraging the \textit{order-preserving} instance code and tracking instance identities without additional matching. 
    }
    \label{fig:teaser}
\end{figure}


In this paper, we propose a robust context fusion method for online VIS. 
To preserve the location information of instances, we build our model on the crop-free transformer-based network. As shown in Figure~\ref{fig:teaser}, we employ an attention-based context fusion module for multi-modal contextual information interaction, in which we compress the reference features to mitigate redundancy, improve robustness, and also introduce audio cues. An order-preserving instance code is further learned by the transformer decoder with a fixed-length instance query. Also, by leveraging the Lipschitz continuity of the network, we employ a matching-free instance identity tracking approach. The instance identity (shown as \textcolor{red}{red} and {\color[RGB]{0,180,52}{green}} colors in Figure~\ref{fig:teaser}) can be directly corresponded to the index of the instance code, thus greatly reducing the matching cost in the inference stage. 
From experimental and mathematical perspectives, we demonstrate that the order-preserving property of instance codes is tight under natural scenes even without any supervision. This can shed light on new directions of instance tracking.
Our context fusion module is flexible to fuse multi-modal signals.
We construct an unconstrained audio-visual dataset for the VIS task, and demonstrate that audio is helpful for the VIS task but the improvement is not significant due to the weak correlation between audio and visual signals in the wild scenarios.
Our contributions are three-fold. 
\begin{itemize}
    \item A robust context fusion module for modeling compact and effective spatio-temporal correlations for online VIS. Our method achieves the state-of-the-art performance among online VIS methods on Youtube-VIS 2019 and 2021 benchmarks.
    \item A matching-free and supervision-free instance identity tracking method, and its corresponding mathematical explanation.
    \item We are the first to investigate audio effects on the video dense prediction task and contribute a corresponding dataset with synchronized audio-visual signals.
\end{itemize}

A preliminary version of our method has been published in \cite{li2021hybrid}. In this manuscript, we make the following improvements. 1) We analyze the mathematical intuition behind the order-preserving property of instance code and conduct more experiments to understand its behavior. 2) We simplify the instance code generation to improve the network generalization, and at the same time the computational cost is reduced. 3) We extend the framework to accept audio-visual inputs, and present an effective method to fuse the multimodal context information leveraging the compatibility of the network. 4) We provide extensive quantitative and
qualitative experimental results to show the performance of our framework in different settings and examine the effectiveness of the key modules in ablation studies.

\section{Related Work}

\begin{figure*}[t]
\centering
\includegraphics[width=\textwidth]{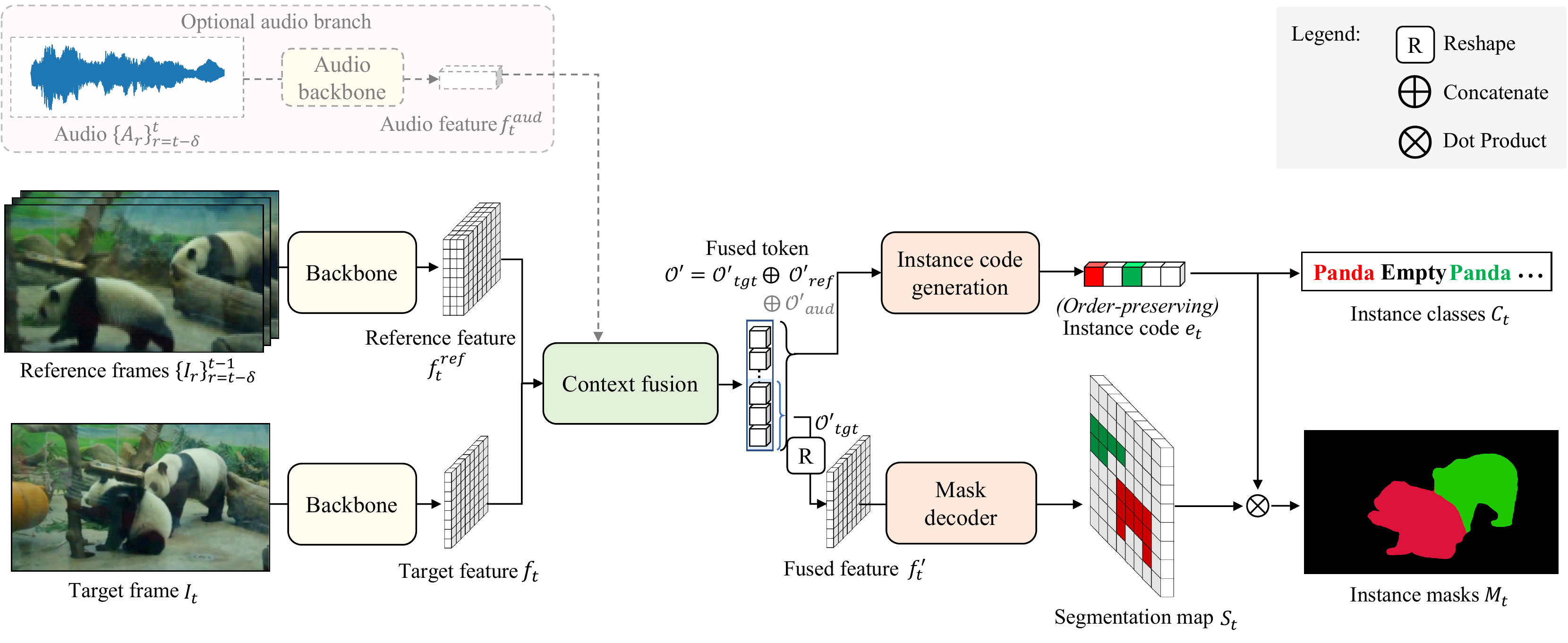}
\caption{\textbf{Overview of the proposed network at timestamp \bm{$t$}.} 
For each target frame $I_t$, we consider reference frames $\{I_r\}_{r=t-\delta}^{t-1}$ as context for predicting the instance masks $M_t$ and classes $C_t$. Both target feature $f_t$ and reference feature $f_t^{ref}$ are extracted by a shared backbone and fused in the context fusion module. The mask decoder decodes the fused target token $\mathcal{O}^{\prime}_{tgt}$ into the segmentation map $S_t$. The transformer decoder decodes the fused token $\mathcal{O^{\prime}}$ into an order-preserving instance embedding $e_t$ in which each slot corresponds to a specific instance in $I_t$.
The final predictions, instance classes $C_t$ and instance masks $M_t$, are obtained from the instance embedding $e_t$ and segmentation map $S_t$ with a simple dot product operation.
For the instance identity matching, we constrain the slot indices (indicated in \textcolor{red}{red} and {\color[RGB]{0,180,52}{green}}) of instance embedding $e_t$ to represent the instance identity. 
The optional audio signal serves as another context and can be fused in the same way as reference frames.
}
\label{fig:overview}
\end{figure*}

\para{Video instance segmentation.}
Video instance segmentation \cite{yang2019video,cao2020sipmask,liu2021sg,tian2019fcos,tian2019fcos,yang2021crossover,li2021video} requires classifying and segmenting each instance in a frame and assigning the same instance with the same identity across frames. 
There are mainly two types of methods for VIS tasks: online and offline.

Online VIS: Given a small set of preceding \textit{reference} frames, online VIS aims to segment, classify and track object instances in each \textit{target} frame. Mask-Track-RCNN \cite{yang2019video} is the first attempt to address the VIS problem in an online setting. It extends the Mask-RCNN \cite{he2017mask} with a tracking head to associate instance identities. SipMask \cite{cao2020sipmask} and SG-Net \cite{liu2021sg} build the tracking head on top of the modified one-stage still-image instance segmentation method FCOS \cite{tian2019fcos} and Blender-Mask \cite{chen2020blendmask}, and achieve better speed and performance compared to MaskTrack-RCNN.
CrossVIS \cite{yang2021crossover} introduces the cross-over learning scheme and global instance embedding to learn better features for robust instance tracking and segmentation. HITF proposes inter-frame attention and intra-frame attention layers which bridges instance information across frames by an instance code.

Offline VIS: The offline methods handle the sequence-to-sequence prediction, where all frames are considered to be equivalent. Given a clip of video, offline VIS is to segment, classify and track object instances in all clip-level frames at the same time. VISTR \cite{wang2021end} utilizes a transformer encoder on top of the convolutional backbone and matches instance identities by transformer decoder. IFC introduces a frame memory to reduce the computational cost for temporal aggregation, which decouples the segmentation in each frame and only communicates via frame memory.


\para{Video object segmentation.}
Video object segmentation (VOS) aims to segment object masks across frames in a class-agnostic manner. Typically, additional cues are given to specify the target object. Semi-supervised VOS \cite{yang2018efficient,jain2017fusionseg,wang2019ranet,caelles2017one,oh2019video} gives the first frame object mask to specify the target object. RANet \cite{wang2019ranet} proposes a ranking attention module to filter our irrelevant features based on the pixel-level similarity obtained from an encoder-decoder network. Space-Time-Memory networks (STM) \cite{oh2019video} introduces a new paradigm that builds a memory bank for each object in the video and segments following objects by matching them to the memory bank. More recently, referring video object segmentation (R-VOS) emerges and attracts more and more attention because of its strong ability to facilitate human-computer interaction. R-VOS \cite{referformer,botach2021mttr,liang2021clawcranenet,urvos,pminet,li2022r} aims to segment object masks throughout the entire video by giving a linguistic expression. URVOS first segments object masks by visual cues then select the referred one by referring expression. MTTR follows this paradigm while enable the multimodal fusion by a transformer encoder to enhance the semantic consensus between linguistic and visual modalities. ReferFormer directly segments the referred object by employing a language-conditioned instance query in the transformer decoder which avoids segmenting irrelevant objects.

\para{Image Instance Segmentation}
Most of image instance segmentation methods adopt either bottom-up \cite{cheng2020panoptic,xie2020polarmask,wang2020solo,wang2020solov2,wang2020axial,yang2019deeperlab,chen2020naive,wang2020max} or top-down \cite{he2017mask,Cao_SipMask_ECCV_2020,liu2018path,huang2019mask} paradigm. For bottom-up methods, several representations can be adopted to represent instance identities, such as object centers \cite{cheng2020panoptic}, object-specific coefficients \cite{Cao_SipMask_ECCV_2020}. PolorMask \cite{xie2020polarmask} directly models instance contour by using 36 uniformly-spaced rays in polar coordinates, which can be assumed as a generalization of box representation that models each instance contour by 4 rays in polar coordinates. SOLO \cite{wang2020solo,wang2020solov2} discriminates each instance by its location and size. Panoptic-Deeplab \cite{cheng2020panoptic} models semantic segmentation, object center heatmap and center offset separately and then assembles those components to instance masks. SipMask \cite{Cao_SipMask_ECCV_2020} follows the previous one-stage method FCOS \cite{tian2019fcos} to represent each instance by object-specific coefficients and corresponding mask prototypes. To achieve more accurate segmentation result, SipMask divides the input image to four parts and predicts masks in each part separately. For top-down methods, instance identity is mainly represented by the bounding-box of detected object. Mask R-CNN \cite{he2017mask} employs a region proposal generation network (RPN) equipped with RoIAlign feature pooling strategy and a feature pyramid networks (FPN) \cite{lin2017feature} to obtain fixed-sized features of each proposal. The pooled features are further used for bounding box prediction and mask segmentation. Followed by Mask R-CNN, several methods are proposed to improve pooling and confidence scoring strategy \cite{li2017fully,liu2018path,huang2019mask}.

\para{Audio-visual representative learning.}
Audio-visual representative learning aims to correspond the sound to their sources in the video frames. Sound localization \cite{zhao2018sound, tian2018audio, senocak2018learning,xu2020cross,mittal2020emotions,cheng2020look} is one of the more extensively explored tasks in this domain, which locates the sound sources of the audio recording in the image. Sound-of-pixel \cite{zhao2018sound} conducts sound source localization and separation simultaneously on an instrument dataset, by estimating time-frequency masks on the audio spectrogram from the video, and recovering the separated sounds through an inverse Short Time Fourier Transform (STFT).
A2V \cite{tian2018audio} leverages the attention mechanism to link the audio and video in unconstrained videos. 
Similarly, Senocak et al. \cite{senocak2018learning} further expand the audio-visual localization task into semi-supervised scenarios and managed to improve the localization performance by introducing unsupervised loss. 
Audio-visual sound source (speaker) separation \cite{afouras2018conversation, ephrat2018looking, lu2018listen, morrone2019face, gabbay2018seeing, zhao2018sound} aims to isolate the target sound in a noisy scene. Recent methods leverage visual cues to guide the separation. Gabbay et al. \cite{gabbay2018seeing} first feed the video frames into a video-to-speech model to predict the speaker's voice from facial movements, then use it to separate the target speaker from sound mixtures.

\section{Method}
\subsection{\bf Overview}
\begin{figure}[t]
\centering
\includegraphics[width=\linewidth]{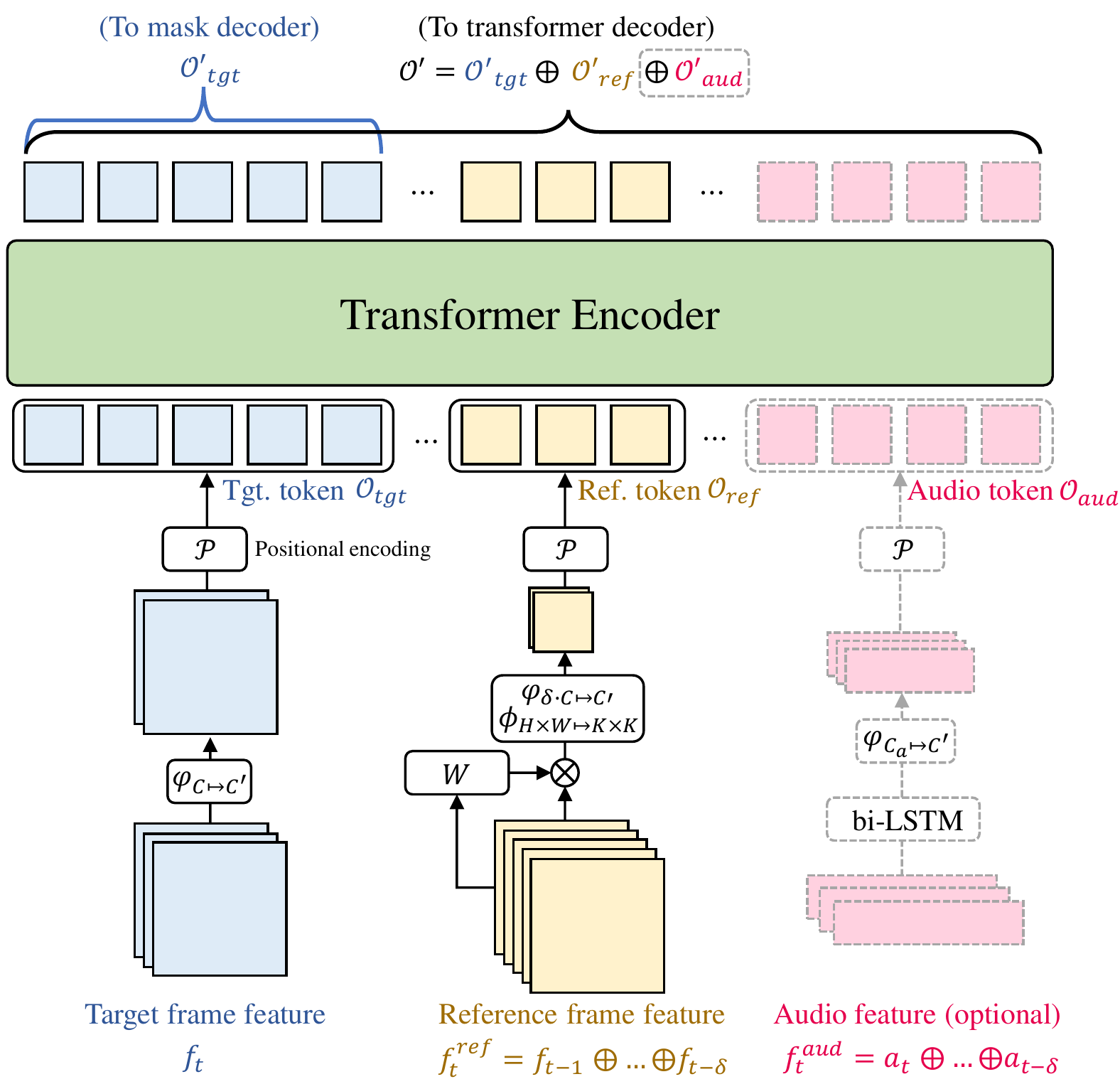}
\caption{\textbf{Robust context fusion module.} The target frame feature $f_t$ is projected to a lower dimension, and then flattened and added with positional encoding to be tokens $\mathcal{O}_{tgt}$. Considering the importance to the target prediction, features from reference frames $f_t^{ref}$ are first reweighted by a learned mask $W$, and then compressed to lower spatial and channel dimensions as $\mathcal{O}_{ref}$. The optional audio features are projected to a lower dimension as $\mathcal{O}_{aud}$ after bi-LSTM. We concatenate the source tokens and feed them to the transformer encoder to model their correlations.}
\label{fig:trE}
\end{figure}

\para{Pipeline overview.}
We first introduce our transformer-based network targeting online video instance segmentation, and then extend it by adding optional synchronized audio signals. 
The pipeline is illustrated in \fig{{\ref{fig:overview}}}. For each iteration $t$ with a target frame $I_t$ and reference frames $\{I_{r}\}_{r=t-\delta}^{t-1}$, we first extract the target feature $f_t$ and reference features $\{f_r\}_{r=t-\delta}^{t-1}$ with a shared backbone. 
Considering the redundancy and noise in the pixel-wise correlations from reference to target frames, we compress the target and reference context as tokens $\mathcal{O}_{tgt}$ and $\mathcal{O}_{ref}$ respectively. 
Then, a transformer encoder in the robust context fusion (RCF) module fuses the original tokens as  $\mathcal{O}'=\mathcal{O}'_{tgt} \oplus \mathcal{O}'_{ref}$, where $\mathcal{O}'_{tgt}$ and $\mathcal{O}'_{ref}$ are fused tokens correspond to target and reference contexts.
Afterwards, $\mathcal{O}'_{tgt}$ is decoded into the target segmentation map $S_t$ and the overall $\mathcal{O}'$ is decoded into an instance code $e_t$, which represents order-preserving instance identities.
We directly compose the final instance segmentation map $M_t$ by dynamic convolution between $S_t$ and $e_t$ without any sophisticated matching algorithm. 
We elaborate our online VIS framework in Section \ref{sec:online-vis}.

Our pipeline is flexible to aggregate multi-modal contexts. Audio is verified to be helpful for object recognition and localization, and we investigate the audio-visual correspondence and leverage it to improve the dense prediction (instance segmentation in this work). 
We align and fuse audio features $f_t^{aud}$ in the RCF the same way as reference video frames.
The online audio-visual instance segmentation framework is detailed in Section \ref{sec:online-avis}.

\subsection{\bf Online Video Instance Segmentation}
\label{sec:online-vis}
Since our method performs online inference,  we process one target frame $I_t$ at each iteration with additional reference frames $\{I_r\}_{r=t-\delta}^{t-1}$. The target features $f_t$ are extracted by a shared backbone, while reference features $\{f_r\}_{t-\delta}^{t-1}$ are obtained from previous iterations.
Note that feature extraction for each frame is performed only once in the online inference stage, thus saving processing time.
The extracted video feature of both target frame and reference frames are sent to the robust context fusion module for further interaction.

\subsubsection{\bf Robust Context Fusion (RCF)}
We fuse the context information using compact and representative visual tokens for target and reference frames by taking their importance into consideration. The structure of RCF is illustrated in \fig{\ref{fig:trE}}.

\para{Target token.}
Since the target frame contains the most important information about spatial and semantic cues, we only compress the feature map along the channel dimension and preserve the spatial dimension.
{
\begin{equation}
    \mathcal{O}_{tgt} = \mathcal{P}(\varphi _{C \mapsto C' }  (f_t)),
\end{equation}}
where $\varphi _{C \mapsto C' }$ is a $1\times 1$ conv layer to project the feature map  $f_t\in\mathbb{R}^{C\times H\times W}$ to a lower dimension $\mathbb{R}^{C^{\prime}\times H\times W}$, $\mathcal{P}$ denotes operations to flatten the feature and add it with positional encoding.

\begin{figure}[t]
\centering
\includegraphics[width=\linewidth]{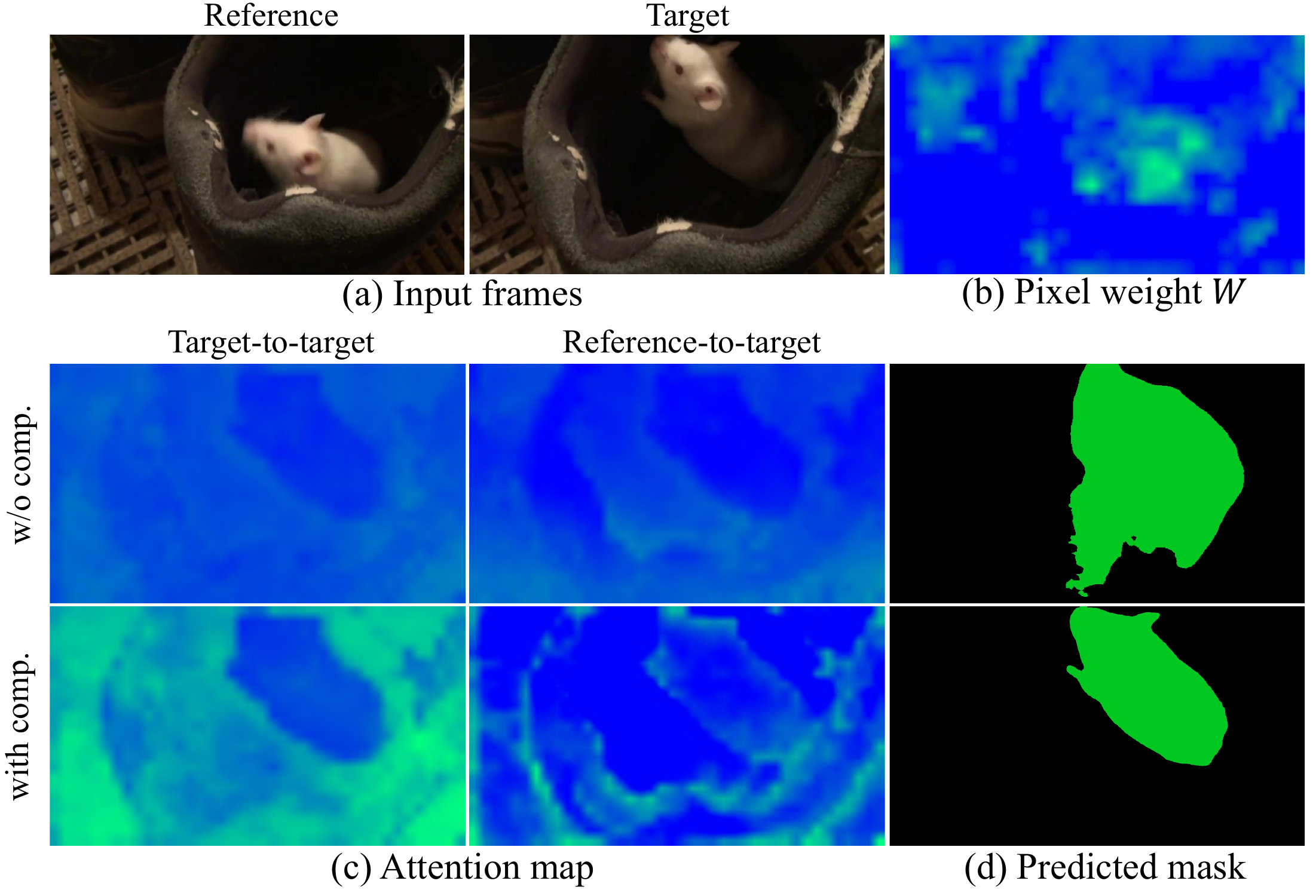}
\caption{
\textbf{Visualization of intermediate components in RCF}. Given the hard cases (a) with confusing background, our design with token compression can generate more representative attention maps (c) and get better result (d).
}
\label{fig:token_compress_illustration}
\end{figure}

\para{Reference token.}
Reference tokens are used to enhance target tokens according to their correlations with the target, while their own representation is less important. We employ compact and representative reference tokens to alleviate matching noise and enhance the importance of the target. Feature compression in both spatial and channel dimensions is applied.
{
\begin{equation}
    \mathcal{O}_{ref} = \mathcal{P} (\varphi _{\delta \cdot C \mapsto C' } (\phi _{H\times W \mapsto K \times K }  ({W}\cdot f_t^{ref}))),
\end{equation}}
where $f_t^{ref} = f_{t-1}\oplus ,...,\oplus f_{t-\delta}$ is a concatenated feature of all reference features, $\phi _{H\times W \mapsto K \times K } $ is a pooling or depth-wise conv layer to compress the spatial dimension from $H\times W$ to $K\times K$, $\varphi _{ \delta \cdot C \mapsto C' }$ is a $1\times 1$ conv layer to compress the channel dimension from $\delta \cdot C$ to $C'$, ${W} = \varphi _{\delta \cdot C \mapsto 1 } (f_{ref})$ is a learned pixel-wise weight map (visualized in \fig{\ref{fig:token_compress_illustration}} (b)). We get the final reference tokens as $\mathcal{O}_{ref} \in \mathbb{R}^{C^{\prime}\times K\cdot K}$.

\para{Token fusion.}
The target tokens $\mathcal{O}_{tgt}$ and reference tokens $\mathcal{O}_{ref}$ are concatenated and fused with a transformer encoder, and then we get the fused tokens $\mathcal{O}' = \mathcal{O}'_{tgt} \oplus  \mathcal{O}'_{ref} \in \mathbb{R}^{C^{\prime}\times (H\cdot W+K\cdot K)}$. The fused target tokens $\mathcal{O}'_{tgt} \in \mathbb{R}^{C^{\prime}\times H\cdot W}$ are further reshaped and fed into a mask decoder to generate the segmentation map $S_t$.
In addition, the whole fused tokens $\mathcal{O}'$ are fed into a transformer decoder to extract instance-specific information, which is discussed in Section \ref{sec:decoder}.



\para{RCF analysis.}
We further analyze our design of RCF from theoretical and experimental perspectives.
Following the transformer structure \cite{bello2019attention}, we adopt standard the scaled dot-product attention, \ie, $\mathrm{Attention}(Q,K,V) = \mathrm{softmax}(\frac{QK^T}{\sqrt{d_k}}) V$, in the transformer encoder of RCF. A softmax function is applied to obtain the weights on the values. Given a much smaller token set (with size $|C^\prime \times K\times K|$) of references, the target tokens ($|C^\prime \times H \times W|$) dominate the attention in the transformer encoder. 
We thus get larger weights for target values of the compressed tokens than the uncompressed counterpart. 
The target-to-target attention map in \fig{\ref{fig:token_compress_illustration}} (c) shows that target tokens in the transformer play more important roles than that without compression. 
Besides, the compressed tokens filter out irrelevant or noisy pixels and contain global information of references frames, thus are more discriminative and generate more representative attention. The reference-to-target attention is visualized in Figure~\ref{fig:token_compress_illustration} (c), where the compressed version looks more informative.
According to the above two factors, our robust token compression can generate a better segmentation mask.

We underline that the goal of our online method is different from the offline methods \cite{wang2021end,hwang2021video}, and the network designs should be different accordingly. The offline methods tackle sequence-to-sequence prediction. The importance of all frames is equivalent, and compressing reference context is not mercenary. For the online setting, the target for each prediction is only the current frame, and importance-aware compression could benefit as discussed above.

\subsubsection{\bf Decoder}
\label{sec:decoder}

\begin{figure}[htbp]
\centering
\includegraphics[width=0.9\linewidth]{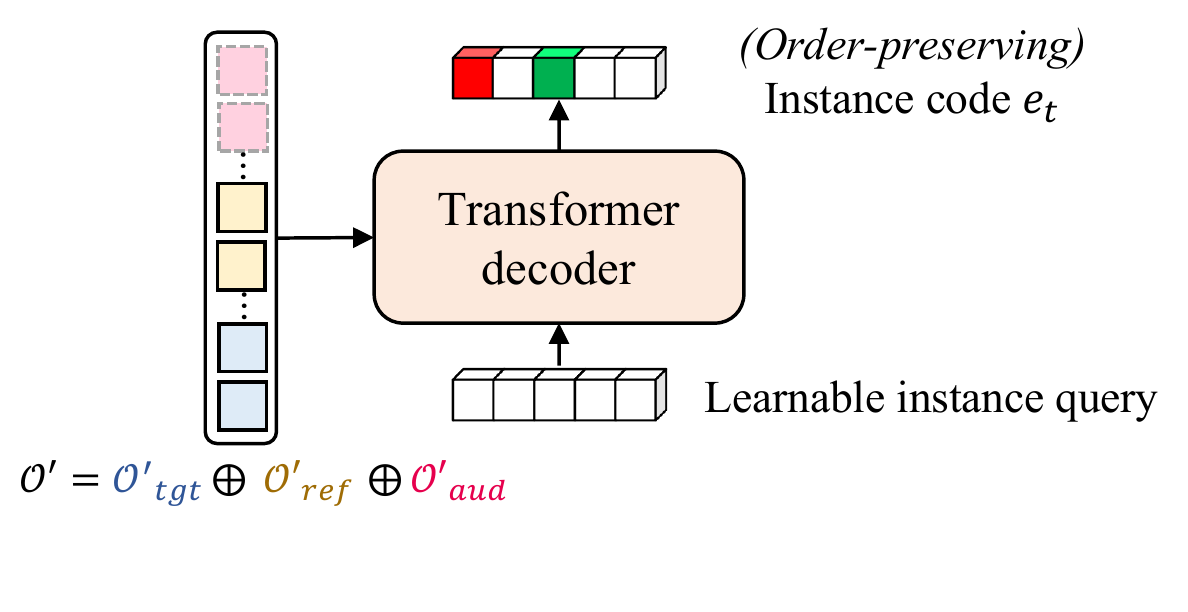}
\vspace{-0.5cm}
\caption{\textbf{Illustration of instance code generation.} A learnable instance query is utilized to query the output of transformer encoder $\mathcal{O}^\prime$ to form the instance code $e_t$.}
\label{fig:code generation}
\end{figure}

We generate instance code using a transformer decoder with a learnable instance query. As shown in Figure~\ref{fig:code generation}, the learnable instance query is used to decode instance-specific information from the fused tokens $\mathcal{O}^\prime$ using a transformer decoder, where instance query is a set of learnable embedding as previous methods \cite{wang2021end,hwang2021video,wu2021seqformer}. We term the decoded instance-specific features as instance code $e_t\in \mathbb{R}^{C_{e}\times N}$, where $C_e$ and $N$ are the channel dimension and length of the instance code respectively. For decoding the segmentation mask, we follow the FPN structure \cite{lin2017feature} to fuse the low-level features. Let the upsampled segmentation map be $S_{t}\in \mathbb{R}^{C_o\times H_o\times W_o}$ where $C_o$, $H_o$ and $W_o$ are the dimension, height and width of the upsampled segmentation map. After that, we leverage dynamic convolution ~\cite{jia2016dynamic} to obtain instance masks $M_t$ from instance code. 
In particular, dynamic filters $\theta_t\in \mathbb{R}^{C_o\times N}$ are learned from instance code by two fully connected layers. The mask prediction $M_t\in \mathbb{R}^{N\times H_o\times W_o}$ can be computed as $M_t = \theta_t^TS_t$.

\subsubsection{\bf Loss Function}
We assign each prediction with a ground-truth label then apply a set of loss functions between them. 
Given a set of predictions ${\{\hat{p}_i(c), \hat{m}_i\}_{i=0}^{N}}$ and a set of ground-truth $\{c_i, m_i\}_{i=0}^{N}$ (padded with $\emptyset$), we search for an assignment $\sigma \in\mathcal{S}_N$ with highest similarity, where $p_i(c)$ is the probability of class $c$ (including $\emptyset$) and $m_i$ is the mask of the $i$-th instance respectively. $\mathcal{S}_N$ is a set of permutations of N elements. The similarity can be computed as 
{
\begin{equation}
    \mathrm{Sim} = \mathbbm{1}_{\{c_i\neq\emptyset\}} \big [\mathrm{Dice}(m_{\sigma(i)}, \hat{m}_i) + \hat{p}_{\sigma(i)}(c_i)\big],
\end{equation}
}
where $\mathbbm{1}$ is an indicator function and $\mathrm{Dice}$ indicates the Dice loss \cite{milletari2016v}. The best assignment $\hat{\sigma}$ is solved by the Hungarian algorithm \cite{kuhn1955hungarian}. Given the best assignment $\hat{\sigma}$, the loss between ground-truth and predictions can be computed as 
{
\begin{equation}
    \mathcal{L} = \sum_{i=0}^{N}-\log\hat{p}_{\hat{\sigma}(i)}(c_i)+\mathbbm{1}_{\{c_i\neq\emptyset\}}\mathrm{Dice}(\hat{m}_{\hat{\sigma}(i)}, m_i),
\end{equation}}
Following \cite{cheng2021per}, we only consider the mask loss when the class prediction is not the empty class.

\subsubsection{\bf Instance Identity Matching}
\label{sec:instance_matching}
Unlike previous online methods \cite{yang2021crossover,cao2020sipmask,yang2019video} cropping out instances and using multiple cues (\eg, class, position and appearance) for matching instances across frames, we directly leverage the constraint that non-empty predictions from the same slot of the instance code have the same identity. 

\para{Order-preserving instance code.}
Our simple instance identity matching approach takes advantage of the Lipchitz continuity of our network (Appendix A) on the condition of bounded inputs. Given the Lipchitz continuity of our network $\Theta(\cdot)$ on the normalized image space $\mathcal{I}$, the relation of input and prediction can be represented as
{
\begin{equation}
    0\leq \|\Theta(I_t) - \Theta(I_{t-1})\|_p\leq \lambda \|I_t - I_{t-1}\|_p,
\end{equation}}
where $\lambda$ is the Lipchitz constant and $\|\cdot\|_p$ represents p-norm with $p\in[1, \infty]$. In most cases, the discrepancy of adjacent frames is small, leading the discrepancy of their outputs $\Theta(\cdot)$ to be also small. Formally,
{
\begin{equation}
\lim\limits_{\|I_t - I_{t-1}\|_p\to 0}\|\Theta(I_t) - \Theta(I_{t-1})\|_p= 0.
\label{equ:lip}
\end{equation}}
Since the order of output $\Theta(I_t)=\{\hat{p}_i(c), \hat{m}_i\}_{i=0}^N$ is directly linked to the order of the instance code $e_t$, if the order of $e_t$ and $e_{t-1}$ are different, the discrepancy $\|(\Theta(I_t) - \Theta(I_{t-1})\|_p > \Delta$ could be large and does not satisfy Equation~\ref{equ:lip}. Therefore, the order of instance code $e_t$ would preserve given $\|I_t - I_{t-1}\|_p\to 0$. We utilize the preserved instance code order to track instance identities across frames. Different from previous offline methods \cite{wang2021end,hwang2021video,wu2021seqformer, li2021hybrid} that leverage losses to supervise the order-preserving, we claim that order-preserving is a property of our deep model and can work well in the online setting (with only adjacent frames) even without any supervision.

When $\|I_t - I_{t-1}\|_p\to 0$ holds, the order will preserve. While the this property of instance code is not constrained as tight as $\|I_t - I_{t-1}\|_p\to 0$ but related to the local smoothness of the network $\Theta$. In other words, with a good local smoothness at a local region of $\Theta$, even if $\|I_t - I_{t-1}\|_p\to 0$ does not hold, the order can still preserve. We provide with a relaxed deduction: If the order of the instance code changes, the Lipschitz continuity of the network will be violated. Let us denote $\mathcal{I}_{sub} \in \mathcal{I}$ as a local region containing input images $I_t$ and $I_{t-1}$, where $\mathcal{I}$ is the normalized input image space. Equation 5 (Lipschitz continuity) can be rewritten as:
$$||\Theta(I_t) - \Theta(I_{t-1})||_p < \lambda(\mathcal{I}_{sub})\cdot||I_t - I_{t-1}||_p$$
where $\lambda(\mathcal{I}_{sub})$ is the Lipschitz constant of the network at $\mathcal{I}_{sub}$. For the \textbf{cases of order changes}, we denote $\epsilon_c=||I_t - I_{t-1}||_p$ as the input discrepancy, and $\Delta_c=||\Theta(I_t) - \Theta(I_{t-1})||_p$ as the output discrepancy. As the order changes, $\Delta_c$ can be very \textbf{large} according to the network design (Line 498-500). Then, the Lipschitz continuity takes the form:
$$
\frac{\Delta_c}{\epsilon_c} < \lambda(\mathcal{I}_{sub})
$$

If the network is well trained, Lipschitz constant $\lambda(\mathcal{I}_{sub})$ at a local region is always small to ensure the network smoothness.
Since $\Delta_c$ is large and $\epsilon_c$ is a constant determined by the data, $\frac{\Delta_c}{\epsilon_c}$ can probably be larger than $\lambda(\mathcal{I}_{sub})$, which violates the Lipschitz continuity.
Therefore, we can conclude that the order preserving property is mainly determined by  $\lambda(\mathcal{I}_{sub})$ which reflects the local smoothness of the network, rather than purely by the input discrepancy $||I_t-I_{t-1}||$.
In the case of $p=1$, when the order changes, $\frac{\Delta_c}{\epsilon_c}\sim O(10^3)$ in Youtube-VIS dataset. Thus, the order can keep when the $\lambda({\mathcal{I}_{sub}}) < O(10^3)$, which is not a tight constraint. 

We empirically verify that the order-preserving matching solves most cases. However, there are still exceptional cases we cannot assume $\|I_t - I_{t-1}\|_p\to 0$ and $\|(\Theta(I_t) - \Theta(I_{t-1})\|_p$ could also be small if tiny objects exist. Therefore, to enhance the model robustness, if the mask in $i$-th slot in time $t$ has the IoU larger than 0.5 with mask in $j$-th slot in time $t-1$, we directly assume they share the same identity without considering the order. Visualization analysis of fired slots in instance code will be discussed in Section~\ref{sec:exp}.

\subsection{\bf Online Audio-Visual Instance Segmentation}
\label{sec:online-avis}

The video and audio data are naturally synchronized and contain the homogeneous semantic and spatial information of the sound source. Compared to video data, audio stores the information in a more compact representation while 2D spatial locations of sound sources in the video frames can still be conveyed by audio cues \cite{zhao2018sound, tian2018audio, senocak2018learning, morrone2019face, pu2017audio}. 

VIS task requires semantic and spatial information of the objects which \textit{de facto} corresponds to the information conveyed in the audio modality. We investigate the benefit of introducing audio data into VIS task in the following.

\begin{table*}[t]
\centering
\scalebox{1.2}{
    \begin{tabular}{l|c|c|c|c|ccccc} 
    \hline
    \hline
    Method & Pipeline & Backbone & FPS & Latency (s) & AP & AP50 & AP75 & AR1 & AR10\\
    \hline
    VisTR \cite{wang2021end} & Offline & ResNet-101 & 43.5 & $>$1.9 & 38.6 & 61.3 & 42.3 & 37.6 & 44.2\\
    MaskProp \cite{bertasius2020classifying} & Offline & ResNet-101 & - & - & 42.5 & - & 45.6 & - & - \\
    IFC \cite{hwang2021video} & Offline & ResNet-101 & 89.4 & 6.4 & 44.6 & 69.2 & 49.5 & 44.0 & 52.1 \\
    Mask2Former \cite{cheng2021mask2former} & Offline & ResNet-101 & - & - & 49.2 & 72.8 & 54.2 & - & - \\
    SeqFormer$^*$ \cite{wu2021seqformer} & Offline & ResNet-101 & - & - & 49.0 & 71.1 &  55.7 & 46.8 & 56.9 \\
    SeqFormer$^*$ \cite{wu2021seqformer} & Offline & Swin-L & - & - & 59.3 &82.1& 66.4 & 51.7 &  64.4 \\
    \hline
    \hline
    MaskTrack R-CNN \cite{yang2019video} & Online & ResNet-101 & 28.6 & 0.2 & 31.9 & 53.7 & 32.3 & 32.5 & 37.7\\
    SipMask++\cite{cao2020sipmask} & Online & ResNet-101 & 27.8 & 0.2 & 35.0 & 56.1 & 35.2 & 36.0 & 41.2\\
    SG-Net \cite{liu2021sg} & Online & ResNet-101 & - & - & 36.3 & 57.1 & 39.6 & 35.9 & 43.0\\
    CrossVIS \cite{yang2021crossover} & Online & ResNet-101 & 35.6 & 0.19 & 36.6 & 57.3 & 39.7 & 36.0 & 42.0\\
    STMask \cite{li2021spatial} & Online & ResNet-101 & 23.4 & 0.21 & 36.8 & 56.8 & 38.0 & 34.8 & 41.8\\
    PCAN \cite{ke2021prototypical} & Online & ResNet-101 & $<$15 & $>$0.23 & 37.6 & 57.2 & 41.3 & 37.2 & 43.9 \\
    HITF \cite{li2021hybrid} & Online & HITF & $<8$ & $>$0.29 & 41.3 & 61.5& 43.5 & 42.7& 47.8 \\
    \hline
    \textbf{Ours} & Online & ResNet-101 & 23.9 & 0.21 & 40.8 & 63.4 & 43.5 & 42.4 & 46.7 \\
    \textbf{Ours} & Online & Swin-B & 9.0 & 0.28 & 44.9 & 66.2 & 48.1 & 44.3 & 48.3 \\
    \textbf{Ours} & Online & Swin-L & 5.3 & 0.36 & 47.6 & 70.3 & 50.8 & 46.1 & 52.3 \\
    \hline
    \hline
    \end{tabular}
}
\caption{Comparison to the state-of-the-art offline and online VIS methods on the \textbf{Youtube-VIS-2019} validation set. The inference speed is measured on a singe NVIDIA V100 GPU with $batchsize=1$. $^*$ indicates that values are imported from a preprint.
}
\label{tab:ytvis2019}
\end{table*}

\subsubsection{\bf Audio-Visual Instance Segmentation Dataset}
To investigate the influence of the audio data on the video-level fine-grained prediction tasks, we collect a novel dataset, AVIS, containing synchronized audio and video clips. The data is collected from publicly available videos with 20 vocal categories overlapped with Youtube-VIS dataset.
AVIS contains 1427 videos with synchronized raw audio recordings, which are further randomly split by 75\% and 25\% for training and validation usage. All videos are annotated with high-quality instance-level labels in the format of Youtube-VIS dataset.

\subsubsection{\bf Audio Feature Extraction}
\begin{figure}[htbp]
\centering
\includegraphics[width=\linewidth]{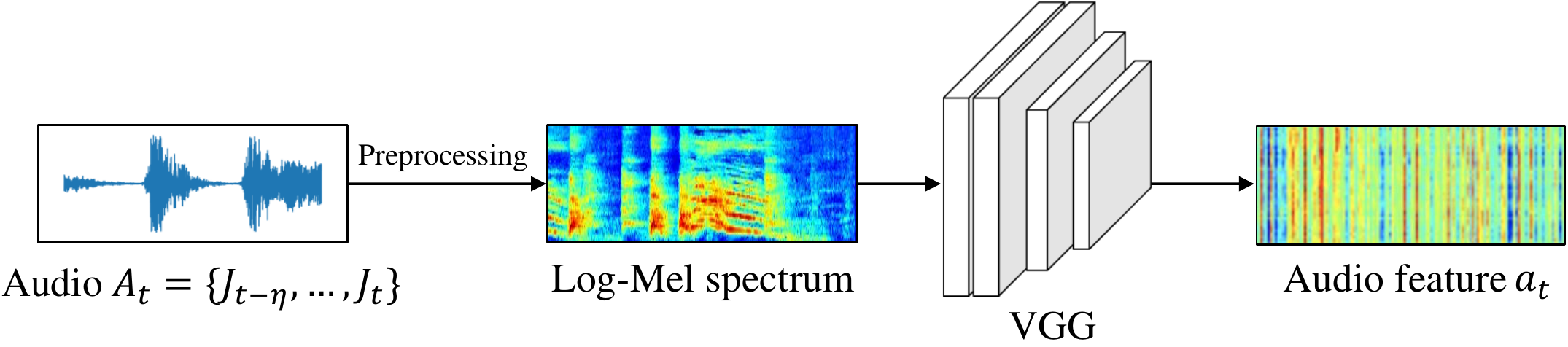}
\caption{\textbf{Audio processing.} The raw audio signal $A_t$ is preprocessed by resampling, STFT, log mel transformation \cite{logan2000mel}, and then fed into the pre-trained VGG \cite{simonyan2014very} model (without the last linear layer) to extract feature $a_t$.
}
\label{fig:audioProcess}
\end{figure}
Since audio data has much higher sampling rate than video data, we combine multiple audio frames to one image. Given the image sequence $\{I_t\}_{t=1}^T$ and audio sequence $\{J_t\}_{t=1}^T$, we combine $A_t=\{J_{t-\eta},\dots,J_{t}\}$ audio frames for each image $I_t$ where $\eta$ is the audio frame length for each image. 
We first resample $A_t$ to 16 kHz mono then compute the spectrum using magnitudes of the Short-Time Fourier Transform (STFT) with a window size of 25 ms, a window hop of 10 ms, and a periodic Hann window. We pad the audio sequence to ensure the output has the same length as the input. The mel spectrogram is computed by mapping the spectrogram to 64 mel bins covering the range 125-7500 Hz. A stabilized log mel spectrogram is further computed by applying log(mel-spectrum + 0.01) where the offset is used to avoid taking a logarithm of zero. Figure~\ref{fig:audioProcess} illustrates the audio feature extraction process. The generated log mel spectrogram \cite{logan2000mel} is fed to the pretrained VGG \cite{simonyan2014very} network to generate audio features. Note that we remove the last linear layer in the VGG to obtain high dimension features.
The extracted audio feature for time $t$ is denoted as $a_t$.

Audio features contain contextual information which relates to the position and semantic categories of the desired instances \cite{zhao2018sound}. However, since audio understanding also depends on the contextual information in audio from prior frames,  we introduce reference audio features as reference video features. Figure~\ref{fig:overview} shows an example of considering $\delta$ reference audio frames. 
The extracted features are sent to the context fusion module for cross-modal fusion.

\begin{figure*}[t]
\centering
\includegraphics[width=\linewidth]{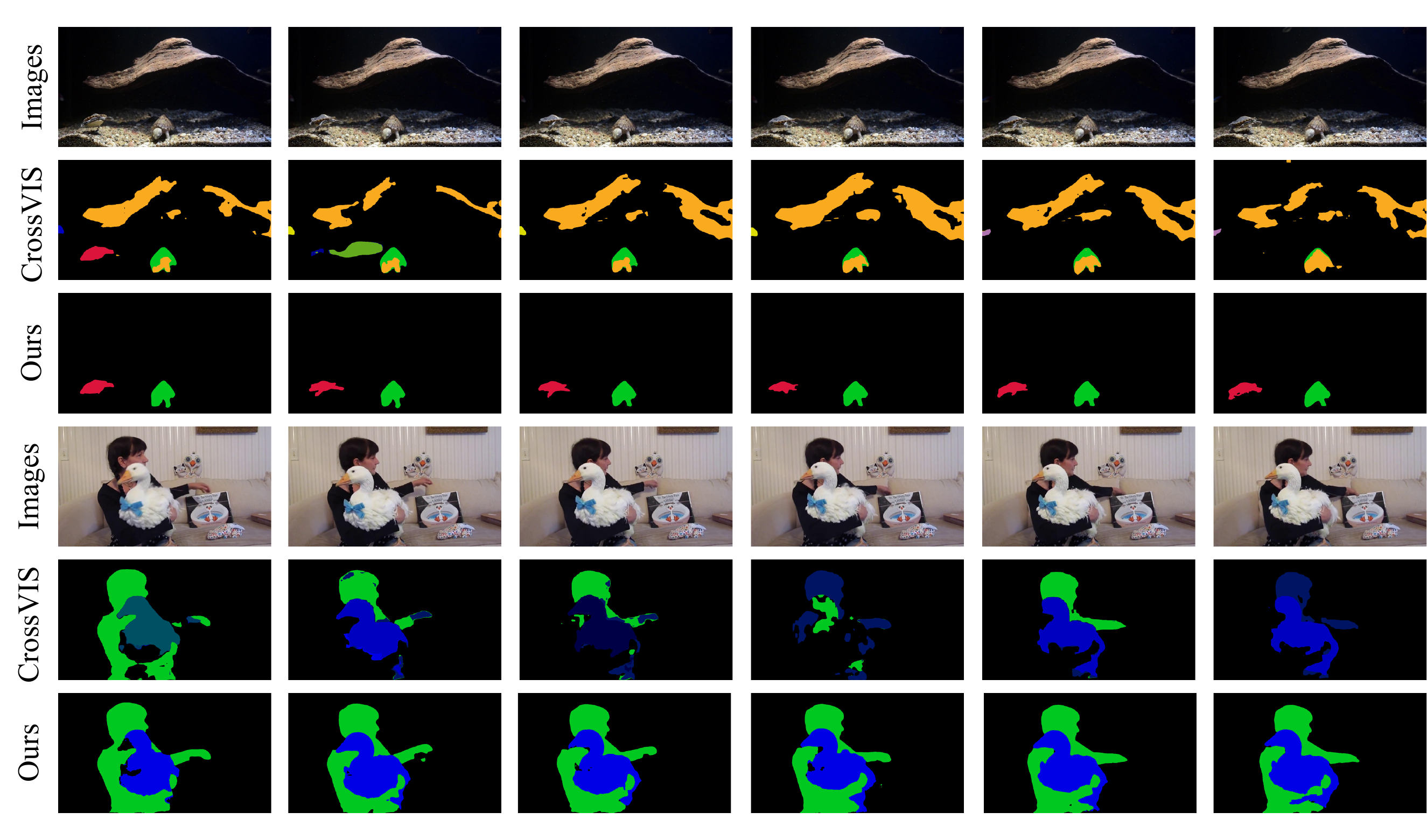}
\caption{Qualitative comparison to the state-of-the-art online video instance segmentation method CrossVIS \cite{yang2021crossover} on the \textbf{Youtube-VIS-2019} validation set.}
\label{fig:vis}
\end{figure*}

\subsubsection{\bf Audio Context Fusion}
We correspond the audio features to pixel-wise visual feature maps with cross-modal attention in the transformer encoder. Similar to tokens from visual features, we create audio tokens to support subsequent context fusion. 

\para{Audio token.}
We combine the overall audio features as the reference audio context $f_t^{aud}=a_{t-\delta} \oplus \cdots \oplus a_t \in \mathbb{R}^{C_a\times (1+\delta)}$, where $C_a=4096$. 
A two-layer bi-directional Long Short Time Memory (bi-LSTM) \cite{hochreiter1997long} network with hidden size 512 is leveraged to aggregate the temporal information. 
After that, we project the audio features $f_t^{aud}$ to a lower dimension $C'$ with two fully-connected layers and ReLU activation. Thus, we get the audio token $\mathcal{O}_{aud} \in \mathbb{R}^{C^{\prime}\times(1+\delta)}$.

\section{Experiment}\label{sec:Experiment}
In this section, we will elaborate on the dataset, implementation details and experiment results for our online VIS and AVIS frameworks.

\subsection{\bf Implementation Details.}
\para{Training.}
We implement our method in the PyTorch framework. Following previous methods \cite{fu2020compfeat, lin2021video}, we first pre-train our model with both Youtube-VIS and overlapped categories on MS-COCO dataset \cite{lin2014microsoft} then finetune the model on the Youtube-VIS dataset. We train our model for 60k iterations with a “poly” learning rate policy with the learning rate $(1-\frac{iter}{iter_{max}})^{0.9}$ for each iteration with an initial learning rate of 0.0006 for all ResNet backbones and 0.0003 for all Swin backbones in experiments. We adopt $\mathrm{batchsize}=16$ and an AdamW \cite{loshchilov2017decoupled} optimizer with $\mathrm{weight\, decay}=10^{-4}$ for all ResNet backbones and $\mathrm{weight\, decay}=10^{-2}$ for all Swin backbones is leveraged. A learning rate multiplier of 0.1 is applied to ResNet backbones, and 1.0 is applied to transformer backbones. Multi-scale training is adopted to obtain a strong baseline. 

\para{Inference.}
The image is resized to $640 \times 360$ during inference. To obtain the final object segmentation result, we first binarize the $M_t\in\mathbb{R}^{N\times H\times W}$ by a threshold of 0.5. Then we filter out slots with a class probability less than 0.4 and keep the remaining ones as the predictions at time $t$. 
\para{Dataset.}
We evaluate our method on the two extensively used VIS datasets, Youtube-VIS-2019 and Youtube-VIS-2021, as well as our newly constructed AVIS dataset.
\begin{itemize}

    \item \textbf{Youtube-VIS-2019} has 40 categories, 4,883 unique video instances. There are 2,238 training videos, 302 validation videos, and 343 test videos in it.
    \item \textbf{Youtube-VIS-2021} is an improved version of Youtube-VIS-2019, which contains 8,171 unique video instances. There are 2,985 training videos, 421 validation videos, and 453 test videos in this dataset. 
    \item \textbf{AVIS} has 20 overlapped categories with Youtube-VIS dataset, 2390 unique video instances.
     There are 1427 videos with synchronized raw audio recordings in it. To the best of our knowledge, AVIS is the first dataset with densely annotated instance-level masks matched to corresponding audio recordings.
\end{itemize}

\para{Metrics.} 
The evaluation metric for this task is defined as the area under the precision-recall curve with different IoUs as thresholds.

\begin{figure}[h!]
    \centering
    \includegraphics[width=\linewidth]{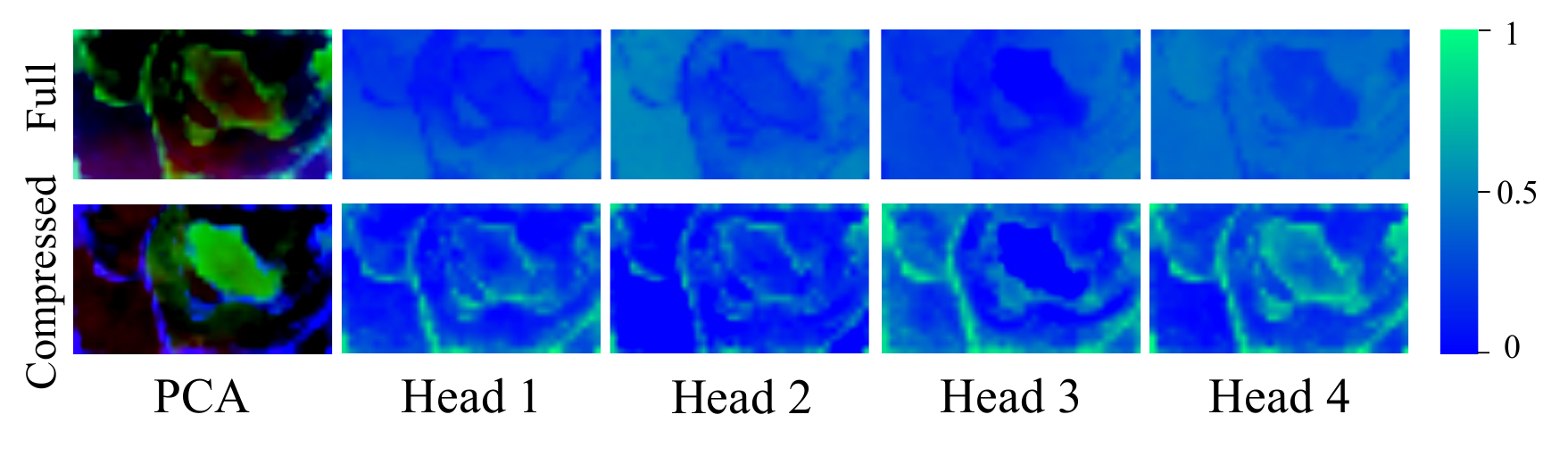}
    \caption{\textbf{Visualization of reference-to-target attention map. The attention maps from compressed tokens are more representative. }  }
    \label{fig:r2t attn}
\end{figure}

\subsection{\bf Video Instance Segmentation Results}
We first present the main results and ablation experiments of VIS task on Youtube-VIS-2019 and Youtube-VIS-2021 dataset without audio involved. 

\subsubsection{\bf Main results}
We compare our method with state-of-the-art methods in this section.

\para{Quantitative result.}
We compare our method against state-of-art VIS methods on \textbf{Youtube-VIS-2019} dataset in Table~\ref{tab:ytvis2019}. (1) Compared to online methods: Our method achieves the best performance of 40.8 mAP when using the same ResNet-101, outperforming the previous start-of-art method PCAN \cite{ke2021prototypical} by a large margin of 3.2 mAP. The recent method HITF \cite{li2021hybrid} leverages a stronger backbone with a higher resolution input, we compare it with our results using the Swin-B backbone for fairness. Our method outperforms HITF by 3.6 mAP with a faster inference speed. (2) Compared to offline methods: Our method eclipses another transformer-based method VisTR \cite{wang2021end} even if it takes video-level input. The results of SeqFormer \cite{wu2021seqformer} is imported from preprint, which uses a stronger deformable-transformer \cite{zhu2020deformable} to conduct temporal fusion, while other settings are similar to VisTR \cite{wang2021end}. Although it achieves an impressive result, we consider it mainly due to the adoption of a stronger deformable transformer. We also compare our method against the state-of-the-art VIS methods on \textbf{Youtube-VIS 2021} in Table~\ref{tab:ytvis2021}. Since Youtube-VIS-2021 is a newly introduced dataset, there are only a few methods for comparison. Our method achieves the best performance among online methods.

\para{Qualitative result.} We compare the qualitative result of our method against CrossVIS \cite{yang2021crossover} on Youtube-VIS-2019 val set in Figure~\ref{fig:vis}. The result indicates that CrossVIS fails to detect or track instances in occluded scenarios, while our methods successfully maintain robust high-quality segmentation and tracking performance. 
More qualitative results are shown in the supplementary material.
As shown in Figure~\ref{fig:r2t attn}, we compare the reference-to-target (R2T) attention map in the transformer encoder using full and compressed reference tokens. The PCA is conducted among all 8 heads of R2T attention map and keeps three main components for RGB channels. We show the first 4 heads of the R2T attention. The attention maps of compressed tokens are sharper in the instance region than those of full tokens, which can also be verified by the PCA map.



\begin{table*}[t]
\centering
\begin{minipage}[t]{0.48\textwidth}
\makeatletter\def\@captype{table}
\centering
\scalebox{1}{
    \begin{tabular}{ccccccc}
    \hline
    \hline
    Method & Backbone & AP & AP\@50 & AP\@75 & AR\@1 & AR\@10 \\
    \hline
    MaskTrack  & ResNet-50 & 28.6 & 48.9 & 29.6 & 26.5 & 33.8\\
    SipMask-VIS & ResNet-50 & 31.7 & 52.5 & 34.0 & 30.8 & 37.8\\
    CrossVIS & ResNet-50 & 34.2 & 54.4 & 37.9 & 30.4 & 38.2\\
    CrossVIS & ResNet-101 & 35.2 & 56.3 & 36.4 & 33.9 & 40.6\\
    HITF & HITF & 35.8 & 56.3 & 39.1 & 33.6 & 40.3 \\
    \textbf{Ours} & ResNet-101 & 37.6 & 56.0 & 41.4 & 35.7 & 42.7 \\
    \hline
    \hline
    \end{tabular}}
    \caption{Comparison to the state-of-the-art online methods on \textbf{Youtube-VIS-2021} validation set.}
    \label{tab:ytvis2021}
\end{minipage}
\hspace{0.01\textwidth}
\begin{minipage}[t]{0.49\textwidth}
\makeatletter\def\@captype{table}
\centering
\scalebox{1}{
    \begin{tabular}{c|ccccc}
    \hline
    \hline
   Ref. Frame & AP & AP\@50 & AP\@75 & AR\@1 & AR\@10\\ 
    \hline
    0 & 37.3 & 58.4 & 40.1 & 37.6 & 43.0\\
    1 & 40.8 & 63.4 & 43.5 & 42.4 & 46.7\\
    2 & 40.4 & 63.8 & 42.3 & 40.5 & 47.0\\
    3 & 40.7 & 64.8 & 44.0 & 39.2 & 45.5\\
    4 & 39.8 & 62.1 & 43.6 & 40.6 & 45.8\\
    5 & 37.2 & 59.5 & 39.1 & 38.4 & 42.5\\
    \hline
    \hline
    \end{tabular}}
    \caption{Impact of the reference frame number on \textbf{Youtube-VIS-2019} validation set.}
    \label{tab:ref_num}
\end{minipage}
\end{table*}
\begin{table*}[h]
\centering
\begin{minipage}[t]{0.49\textwidth}
\makeatletter\def\@captype{table}
\centering
\scalebox{1}{
    \begin{tabular}{cccccc}
    \hline
    \hline
    Method & AP & AP\@50 & AP\@75 & AR\@1 & AR\@10 \\
    \hline
    full-feature Enc. \cite{wang2021end} & 35.6 & 58.8 & 37.3 & 36.6 & 41.3\\
    full-feature Dec. \cite{wang2021end} & 36.6 & 57.7 & 38.0 & 36.9 & 41.8\\
    \textbf{reference-token} & 40.8 & 63.4 & 43.5 & 42.4 & 46.7 \\
    \hline
    \hline
    \end{tabular}}
    \caption{Comparison of different temporal fusion methods on \textbf{Youtube-VIS-2019} validation set.}
    \label{tab:fuse}
\end{minipage}
\hspace{0.01\textwidth}
\begin{minipage}[t]{0.49\textwidth}
\makeatletter\def\@captype{table}
\centering
\scalebox{1}{
    \begin{tabular}{c|ccccc}
    \hline
    \hline
    Token Size & AP & AP\@50 & AP\@75 & AR\@1 & AR\@10\\ 
    \hline
    1 & 38.3 & 58.3 & 39.8 & 38.0 & 43.1\\
    4 & 40.8 & 63.4 & 43.5 & 42.4 & 46.7 \\
    16 & 37.8 & 59.4 & 38.8 & 37.0 & 42.9\\
    64 & 36.3 & 58.5 & 37.9 & 36.4 & 42.0\\
    \hline
    \hline
    \end{tabular}}
    \caption{Impact of the reference-token size on \textbf{Youtube-VIS-2019} validation set.}
    \label{tab:token size}
\end{minipage}
\end{table*}

\begin{table*}[h]
\centering
\begin{minipage}[t]{0.49\textwidth}
\makeatletter\def\@captype{table}
\centering
\scalebox{1}{
    \begin{tabular}{cccccc}
    \hline
    \hline
    Inst. Queries & AP & AP\@50 & AP\@75 & AR\@1 & AR\@10 \\
    \hline
    10 & 40.8 & 63.4 & 43.5 & 42.4 & 46.7 \\
    15 & 40.4 & 63.9 & 44.2 & 40.3 & 45.3\\
    20 & 40.5 & 64.6 & 43.6 & 41.2 & 46.8\\
    25 & 40.3 & 63.3 & 44.2 & 40.4 & 46.1\\
    \hline
    \hline
    \end{tabular}}
    \caption{Impact of number of instance queries on \textbf{Youtube-VIS-2019} validation set.}
    \label{tab:query}
\end{minipage}
\hspace{0.01\textwidth}
\begin{minipage}[t]{0.49\textwidth}
\makeatletter\def\@captype{table}
\centering
\scalebox{1}{
    \begin{tabular}{c|ccccc}
    \hline
    \hline
    Supervision & AP & AP50 & AP75 & AR1 & AR10\\
    \hline
    \Checkmark & 40.5 & 63.5 & 42.6 & 40.5 & 46.7\\
    \XSolidBrush & 40.8 & 63.4 & 43.5 & 42.4 & 46.7\\
    \hline
    \hline
    \end{tabular}}
    \caption{Impact of supervision on the order-preserving property.}
    \label{tab:supervision}
\end{minipage}
\end{table*}

\begin{table}[htbp]
    \centering
    \begin{tabular}{c|ccccc}
    \hline
    \hline
    Weight $W$ & AP & AP50 & AP75 & AR1 & AR10\\
    \hline
    \XSolidBrush & 40.1 & 63.2 & 42.1 & 40.3 & 46.3\\
    \Checkmark & 40.8 & 63.4 & 43.5 & 42.4 & 46.7\\
    \hline
    \hline
    \end{tabular}
    \caption{Impact of pixel-wise weight $W$ on the token fusion.}
    \label{tab:weight}
\end{table}

\para{Inference time.}
Online VIS is mainly for streaming applications. In the video streaming pipeline, the input video is received frame by frame, and latency (including video streaming and model processing time) is the essential time measurement. We further discuss the latency of both online and offline methods for fair comparison. Let $t_{stream}=1/FPS_{stream}$ denote the streaming time of each frame and $t_{model}=1/FPS_{model}$ denote the average model processing time for each frame.
For online methods, the final latency of each frame is $$Latency_{online} = t_{stream} + t_{model}$$
For offline methods, we need to wait till the $N_f$-th frame of the clip comes, where $N_f$ is the frame number of the processed clip. The latency of the last frame (the whole clip) is $$Latency_{offline} = t_{stream}*N_{f} + t_{model}*N_{f}$$
As the videos in Youtube-VIS dataset is of 6 FPS, the latency of our method is 0.171 s, while the latency of the offline method IFC with a clip containing $N_f=36$ frames is 6.4 s ($FPS_{model}=89.4$ reported in [13]). 

As shown in Table~\ref{tab:ytvis2019}, we compare both FPS and latency of our method and previous methods. Our method achieves the best trade-off of accuracy and latency among online methods.

\begin{figure}[htbp]
\centering
\includegraphics[width=1.1\linewidth]{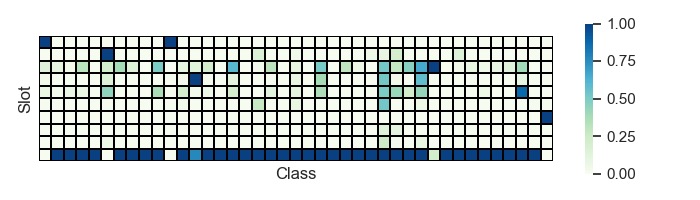}
\caption{\textbf{Illustration of fired slots and class correlation of instance code.} The heatmap is normalized by each class.}
\label{fig:correlation}
\end{figure}

\label{sec:exp}
\para{Analysis of fired slot and class correlation of instance code.} We consider the slot is fired it has a class probability larger than 0.4 and only keep predictions in fired slots as final output. As shown in Figure~\ref{fig:correlation}, we visualize the correlation of fired slots in instance code and each class. We noticed that most of objects are predicted from the same slot while, for several classes, \eg, person, duck and earless seal, they are predicted in stand-alone slots. Those classes are either most commonly appeared or challenging classes in the dataset. In this way, we consider the network encodes some class-specific information in the slot to improve instance segmentation quality.

\subsubsection{\bf Ablation Experiments}
\label{sec:ablation_study}
We conduct ablation studies on Youtube-VIS-2019 to show the effectiveness of different components of our method.

\para{Fusion method.}
To investigate the importance of compressing reference features, we compare the performance of using reference tokens against using two baseline settings that directly fuse full reference features in transformer encoder and transformer decoder respectively. To directly fuse features in transformer encoder, we compute reference tokens similarly to the target token. To fuse features in transformer decoder, we extract reference tokens $\mathcal{O}_{ref}^{\prime}$ by separate transformer encoders and concatenate them before the transformer decoder. As shown in Table~\ref{tab:fuse}, both baseline settings will result in a plummet in performance. Two reasons may account for the decrease. First, the instance position and appearance in the target frame differ from those in reference frames, which can mislead the network if giving reference frames the same importance as the target frame. Second, in transformer encoder, the attention of each layer is normalized by a softmax among all inputs. However, the mask prediction only needs fine-grained target frame information. Therefore, the softmax function downplays the importance of the target frame and introduces noises from reference frames. 

\para{Reference token size.}
The optimal reference token size is a trade-off between information compression loss and target importance gain. We conduct experiments on different token sizes. As shown in Table~\ref{tab:token size}, the token size of 4 achieves the best performance.

\begin{figure*}[t]
\centering
\includegraphics[width=\linewidth]{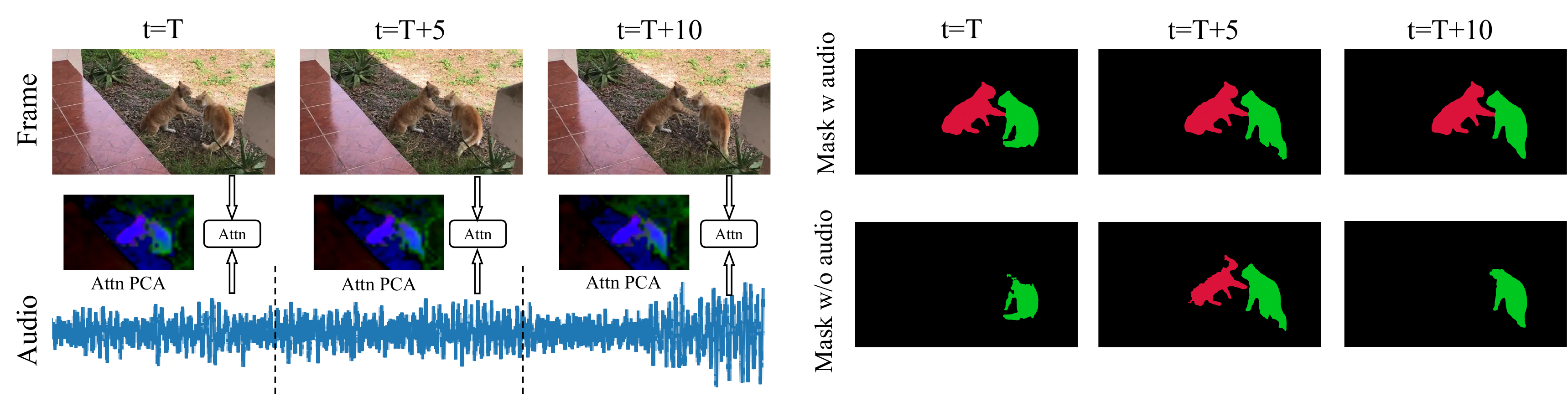}
\caption{Qualitative comparison of the performance with and w/o audio inputs. The multi-head attention map is compressed to three dimension by PCA for visualization.}
\label{fig:av_result}
\end{figure*}

\para{Reference frame number.}
To investigate the importance of temporal information, we conduct an ablation study on training with different reference frames. As shown in Table~\ref{tab:ref_num}, as the reference frame number varies from 0 to 4, the mAP first increases to 40.8 then saturates and decreases. This is because the frames in Youtube-VIS dataset are provided at 6 FPS, thus introducing long-term references may bring in irrelevant information.

\para{Number of instance queries.}
The number of instance queries indicates the maximum detected instances in a frame. However, there may be a varying number of instances in a frame within a video clip. As shown in Table~\ref{tab:query}, we notice that the model is robust to different instance query numbers even when they are severely redundant. The robustness to redundant queries enables the model to stay effective in scenarios that have extremely few instances.

\para{Supervision on the order of instance code.}
Previous online VIS method HITF \cite{li2021hybrid} explicitly gives supervision to keep the order-preserving. However, as we proved, the order-preserving is actually a property of the network. As shown in Table~\ref{tab:supervision}, we notice that the supervision of instance orders cannot improve the performance. 

\para{Pixel-wise weight $W$ for the token fusion.}
Pixel-wise weight $W$ aims to help the reference feature focus on the foreground area. As shown in Table~\ref{tab:weight}, we ablate on the influence of the pixel-wise weight $W$ on the final segmentation performance. We notice that the performance drops 0.7 mAP if we disable the weight in the token fusion.

\subsection{\bf Audio-Visual Instance Segmentation Results}
\begin{table}[h]
\centering
\begin{minipage}[t]{0.5\textwidth}
\makeatletter\def\@captype{table}
\centering
\scalebox{1}{
    \begin{tabular}{ccccccc}
    \hline
    \hline
    Method & Backbone &AP & AP\@50 & AP\@75 & AR\@1 & AR\@10 \\
    \hline
    CrossVIS & ResNet-50 & 42.2 & 59.1 & 47.4 & 42.1 & 49.9 \\
    CrossVIS & ResNet-101 & 44.7 & 60.4 & 50.5 & 42.0 & 49.7 \\
    \textbf{Ours} & ResNet-50 & 46.6 & 63.3 & 51.3 & 44.2 & 51.5 \\
    \textbf{Ours} & ResNet-101 & 48.6 & 63.9 & 52.5 & 44.6 & 52.9 \\
    \hline
    \hline
    \end{tabular}}
    \caption{Comparison to state-of-the-art video instance segmentation on \textbf{AVIS} val set.}
    \label{tab:avis main}
\end{minipage}
\end{table}

We report the performance of using ResNet-50 and ResNet-101 backbone on AVIS dataset. As shown in Table~\ref{tab:avis main}, we also compare the CrossVIS baseline which only leverages the video data. Our method outperforms CrossVIS baseline by 4.4 and 3.9 mAP with ResNet-50 and ResNet-101 backbone respectively.

\begin{table}[t]
    \centering
    \scalebox{1}{
    \begin{tabular}{c|cc|cc|c}
    \hline
    \hline
    Ref. number & AP & AP50 & AP & AP50 & p-value \\
    \hline
    & \multicolumn{2}{c|}{w/o Audio-token} & \multicolumn{2}{c|}{with Audio-token}\\
    \hline
    1 & 42.9 & 60.2 & 44.8 (+1.9) & 60.3 & 0.30 \\
    2 & 44.3 & 60.9 & 46.0 (+1.7) & 63.8 & 0.26 \\
    3 & 45.6 & 61.9 & 46.6 (+1.0) & 63.3 & 0.65 \\
    4 & 45.6 & 62.2 & 45.7 (+0.1) & 62.7 & -\\
    5 & 45.2 & 62.9 & 45.0 (-0.2) & 62.6 & -\\
    \hline
    \hline
    \end{tabular}}
    \caption{Impact of integrating audio-token with different reference frames on \textbf{AVIS} validation set. }
    \label{tab:avis}
\end{table}

To explore the benefit of introducing audio modality to VIS task, we conduct experiments on AVIS dataset to compare the models trained with audio inputs and without audio inputs. 
Figure~\ref{fig:av_result} shows an example where audio improves the segmentation results. The attention between audio and video modality manages to locate the instance which is weakly distinguishable against the background. As shown in Table~\ref{tab:avis}, we notice a slight gain from audio with a small reference frame number. However, as the reference frame number increases, the gain begins to saturate. There are several reasons accounting for the marginal gain. 
For example, there are some sound events that do not last for the whole video duration. Thus, the audio fusion is meaningless for several time steps. 
Moreover, since we add audio data serially with fixed reference audio frames, it is hard to construct a long-term correlation between audios thus the sound event cannot be fully utilized.
To explore the statistical robustness of the gain, we conduct a t-test between corresponding experiments. As shown in Table~\ref{tab:avis}, the p-value indicates the gains are not statistically significant. Therefore, we can safely conclude that while the audio may potentially benefit the video segmentation task, in the current setting, their effect in unconstrained videos is limited.

\section{Conclusion}
\label{sec:Conclusion}
We propose a robust context fusion module for the online VIS task, which corresponds and fuses compact and effective reference features to the target features in a transformer encoder. We observe that the importance-aware compression for reference context is critical in the online setting because the impact of frames are different for the target prediction.
In addition, we leverage an order-preserving instance code to track instance identities, thus avoiding time-consuming matching algorithms. The mathematical explanation indicates that order-preserving is the natural property of the network and can work even without any explicit supervision, which may shed light on new directions of instance tracking.
Our pipeline is flexible and permits multi-modal data. The benefits of using audio context to VIS are seen to be limited in the online setting due to weak correlation between audio and visual data in the wild scenes.


\ifCLASSOPTIONcaptionsoff
  \newpage
\fi

{\small
\bibliographystyle{ieee_fullname}
\bibliography{src/reference}

\begin{thebibliography}{10}\itemsep=-1pt

\bibitem{afouras2018conversation}
Triantafyllos Afouras, Joon~Son Chung, and Andrew Zisserman.
\newblock The conversation: Deep audio-visual speech enhancement.
\newblock {\em arXiv preprint arXiv:1804.04121}, 2018.

\bibitem{athar2020stem}
Ali Athar, Sabarinath Mahadevan, Aljosa Osep, Laura Leal-Taix{\'e}, and Bastian
  Leibe.
\newblock Stem-seg: Spatio-temporal embeddings for instance segmentation in
  videos.
\newblock In {\em European Conference on Computer Vision}, pages 158--177.
  Springer, 2020.

\bibitem{bello2019attention}
Irwan Bello, Barret Zoph, Ashish Vaswani, Jonathon Shlens, and Quoc~V Le.
\newblock Attention augmented convolutional networks.
\newblock In {\em Proceedings of the IEEE/CVF international conference on
  computer vision}, pages 3286--3295, 2019.

\bibitem{bertasius2020classifying}
Gedas Bertasius and Lorenzo Torresani.
\newblock Classifying, segmenting, and tracking object instances in video with
  mask propagation.
\newblock In {\em Proceedings of the IEEE/CVF Conference on Computer Vision and
  Pattern Recognition}, pages 9739--9748, 2020.

\bibitem{botach2021mttr}
Adam Botach, Evgenii Zheltonozhskii, and Chaim Baskin.
\newblock End-to-end referring video object segmentation with multimodal
  transformers.
\newblock {\em arXiv preprint arXiv:2111.14821}, 2021.

\bibitem{caelles2017one}
Sergi Caelles, Kevis-Kokitsi Maninis, Jordi Pont-Tuset, Laura Leal-Taix{\'e},
  Daniel Cremers, and Luc Van~Gool.
\newblock One-shot video object segmentation.
\newblock In {\em Proceedings of the IEEE conference on computer vision and
  pattern recognition}, pages 221--230, 2017.

\bibitem{cao2020sipmask}
Jiale Cao, Rao~Muhammad Anwer, Hisham Cholakkal, Fahad~Shahbaz Khan, Yanwei
  Pang, and Ling Shao.
\newblock Sipmask: Spatial information preservation for fast image and video
  instance segmentation.
\newblock In {\em Computer Vision--ECCV 2020: 16th European Conference,
  Glasgow, UK, August 23--28, 2020, Proceedings, Part XIV 16}, pages 1--18.
  Springer, 2020.

\bibitem{Cao_SipMask_ECCV_2020}
Jiale Cao, Rao~Muhammad Anwer, Hisham Cholakkal, Fahad~Shahbaz Khan, Yanwei
  Pang, and Ling Shao.
\newblock Sipmask: Spatial information preservation for fast instance
  segmentation, 2020.

\bibitem{chen2020blendmask}
Hao Chen, Kunyang Sun, Zhi Tian, Chunhua Shen, Yongming Huang, and Youliang
  Yan.
\newblock Blendmask: Top-down meets bottom-up for instance segmentation.
\newblock In {\em Proceedings of the IEEE/CVF conference on computer vision and
  pattern recognition}, pages 8573--8581, 2020.

\bibitem{chen2020naive}
Liang-Chieh Chen, Raphael~Gontijo Lopes, Bowen Cheng, Maxwell~D Collins, Ekin~D
  Cubuk, Barret Zoph, Hartwig Adam, and Jonathon Shlens.
\newblock Naive-student: Leveraging semi-supervised learning in video sequences
  for urban scene segmentation.
\newblock In {\em European Conference on Computer Vision}, pages 695--714.
  Springer, 2020.

\bibitem{cheng2021mask2former}
Bowen Cheng, Anwesa Choudhuri, Ishan Misra, Alexander Kirillov, Rohit Girdhar,
  and Alexander~G Schwing.
\newblock Mask2former for video instance segmentation.
\newblock {\em arXiv preprint arXiv:2112.10764}, 2021.

\bibitem{cheng2020panoptic}
Bowen Cheng, Maxwell~D Collins, Yukun Zhu, Ting Liu, Thomas~S Huang, Hartwig
  Adam, and Liang-Chieh Chen.
\newblock Panoptic-deeplab: A simple, strong, and fast baseline for bottom-up
  panoptic segmentation.
\newblock In {\em Proceedings of the IEEE/CVF Conference on Computer Vision and
  Pattern Recognition}, pages 12475--12485, 2020.

\bibitem{cheng2021per}
Bowen Cheng, Alexander~G Schwing, and Alexander Kirillov.
\newblock Per-pixel classification is not all you need for semantic
  segmentation.
\newblock {\em arXiv preprint arXiv:2107.06278}, 2021.

\bibitem{cheng2020look}
Ying Cheng, Ruize Wang, Zhihao Pan, Rui Feng, and Yuejie Zhang.
\newblock Look, listen, and attend: Co-attention network for self-supervised
  audio-visual representation learning.
\newblock In {\em Proceedings of the 28th ACM International Conference on
  Multimedia}, pages 3884--3892, 2020.

\bibitem{pminet}
Zihan Ding, Tianrui Hui, Shaofei Huang, Si Liu, Xuan Luo, Junshi Huang, and
  Xiaoming Wei.
\newblock Progressive multimodal interaction network for referring video object
  segmentation.
\newblock {\em The 3rd Large-scale Video Object Segmentation Challenge},
  page~7, 2021.

\bibitem{ephrat2018looking}
Ariel Ephrat, Inbar Mosseri, Oran Lang, Tali Dekel, Kevin Wilson, Avinatan
  Hassidim, William~T Freeman, and Michael Rubinstein.
\newblock Looking to listen at the cocktail party: A speaker-independent
  audio-visual model for speech separation.
\newblock {\em arXiv preprint arXiv:1804.03619}, 2018.

\bibitem{federer2014geometric}
Herbert Federer.
\newblock {\em Geometric measure theory}.
\newblock Springer, 2014.

\bibitem{fu2020compfeat}
Yang Fu, Linjie Yang, Ding Liu, Thomas~S Huang, and Humphrey Shi.
\newblock Compfeat: Comprehensive feature aggregation for video instance
  segmentation.
\newblock {\em arXiv preprint arXiv:2012.03400}, 2020.

\bibitem{gabbay2018seeing}
Aviv Gabbay, Ariel Ephrat, Tavi Halperin, and Shmuel Peleg.
\newblock Seeing through noise: Visually driven speaker separation and
  enhancement.
\newblock In {\em 2018 IEEE International Conference on Acoustics, Speech and
  Signal Processing (ICASSP)}, pages 3051--3055. IEEE, 2018.

\bibitem{he2017mask}
Kaiming He, Georgia Gkioxari, Piotr Doll{\'a}r, and Ross Girshick.
\newblock Mask r-cnn.
\newblock In {\em Proceedings of the IEEE international conference on computer
  vision}, pages 2961--2969, 2017.

\bibitem{hochreiter1997long}
Sepp Hochreiter and J{\"u}rgen Schmidhuber.
\newblock Long short-term memory.
\newblock {\em Neural computation}, 9(8):1735--1780, 1997.

\bibitem{huang2019mask}
Zhaojin Huang, Lichao Huang, Yongchao Gong, Chang Huang, and Xinggang Wang.
\newblock Mask scoring r-cnn.
\newblock In {\em Proceedings of the IEEE/CVF Conference on Computer Vision and
  Pattern Recognition}, pages 6409--6418, 2019.

\bibitem{hwang2021video}
Sukjun Hwang, Miran Heo, Seoung~Wug Oh, and Seon~Joo Kim.
\newblock Video instance segmentation using inter-frame communication
  transformers.
\newblock {\em arXiv preprint arXiv:2106.03299}, 2021.

\bibitem{jain2017fusionseg}
Suyog~Dutt Jain, Bo Xiong, and Kristen Grauman.
\newblock Fusionseg: Learning to combine motion and appearance for fully
  automatic segmentation of generic objects in videos.
\newblock In {\em 2017 IEEE conference on computer vision and pattern
  recognition (CVPR)}, pages 2117--2126. IEEE, 2017.

\bibitem{jia2016dynamic}
Xu Jia, Bert De~Brabandere, Tinne Tuytelaars, and Luc~V Gool.
\newblock Dynamic filter networks.
\newblock {\em Advances in neural information processing systems}, 29:667--675,
  2016.

\bibitem{ke2021prototypical}
Lei Ke, Xia Li, Martin Danelljan, Yu-Wing Tai, Chi-Keung Tang, and Fisher Yu.
\newblock Prototypical cross-attention networks for multiple object tracking
  and segmentation.
\newblock {\em arXiv preprint arXiv:2106.11958}, 2021.

\bibitem{kim2021lipschitz}
Hyunjik Kim, George Papamakarios, and Mnih.
\newblock The lipschitz constant of self-attention.
\newblock In {\em International Conference on Machine Learning}, pages
  5562--5571. PMLR, 2021.

\bibitem{kuhn1955hungarian}
Harold~W Kuhn.
\newblock The hungarian method for the assignment problem.
\newblock {\em Naval research logistics quarterly}, 2(1-2):83--97, 1955.

\bibitem{li2021spatial}
Minghan Li, Shuai Li, Lida Li, and Lei Zhang.
\newblock Spatial feature calibration and temporal fusion for effective
  one-stage video instance segmentation.
\newblock In {\em Proceedings of the IEEE/CVF Conference on Computer Vision and
  Pattern Recognition}, pages 11215--11224, 2021.

\bibitem{li2021hybrid}
Xiang Li, Jinglu Wang, Xiao Li, and Yan Lu.
\newblock Hybrid instance-aware temporal fusion for online video instance
  segmentation.
\newblock {\em arXiv preprint arXiv:2112.01695}, 2021.

\bibitem{li2021video}
Xiang Li, Jinglu Wang, Xiao Li, and Yan Lu.
\newblock Video instance segmentation by instance flow assembly.
\newblock {\em arXiv preprint arXiv:2110.10599}, 2021.

\bibitem{li2022r}
Xiang Li, Jinglu Wang, Xiaohao Xu, Xiao Li, Yan Lu, and Bhiksha Raj.
\newblock R\^{2}vos: Robust referring video object segmentation via relational
  multimodal cycle consistency.
\newblock {\em arXiv e-prints}, pages arXiv--2207, 2022.

\bibitem{li2017fully}
Yi Li, Haozhi Qi, Jifeng Dai, Xiangyang Ji, and Yichen Wei.
\newblock Fully convolutional instance-aware semantic segmentation.
\newblock In {\em Proceedings of the IEEE conference on computer vision and
  pattern recognition}, pages 2359--2367, 2017.

\bibitem{liang2021clawcranenet}
Chen Liang, Yu Wu, Yawei Luo, and Yi Yang.
\newblock Clawcranenet: Leveraging object-level relation for text-based video
  segmentation.
\newblock {\em arXiv preprint arXiv:2103.10702}, 2021.

\bibitem{lin2021video}
Huaijia Lin, Ruizheng Wu, Shu Liu, Jiangbo Lu, and Jiaya Jia.
\newblock Video instance segmentation with a propose-reduce paradigm.
\newblock {\em arXiv preprint arXiv:2103.13746}, 2021.

\bibitem{lin2017feature}
Tsung-Yi Lin, Piotr Doll{\'a}r, Ross Girshick, Kaiming He, Bharath Hariharan,
  and Serge Belongie.
\newblock Feature pyramid networks for object detection.
\newblock In {\em Proceedings of the IEEE conference on computer vision and
  pattern recognition}, pages 2117--2125, 2017.

\bibitem{lin2014microsoft}
Tsung-Yi Lin, Michael Maire, Serge Belongie, James Hays, Pietro Perona, Deva
  Ramanan, Piotr Doll{\'a}r, and C~Lawrence Zitnick.
\newblock Microsoft coco: Common objects in context.
\newblock In {\em European conference on computer vision}, pages 740--755.
  Springer, 2014.

\bibitem{liu2021sg}
Dongfang Liu, Yiming Cui, Wenbo Tan, and Yingjie Chen.
\newblock Sg-net: Spatial granularity network for one-stage video instance
  segmentation.
\newblock In {\em Proceedings of the IEEE/CVF Conference on Computer Vision and
  Pattern Recognition}, pages 9816--9825, 2021.

\bibitem{liu2018path}
Shu Liu, Lu Qi, Haifang Qin, Jianping Shi, and Jiaya Jia.
\newblock Path aggregation network for instance segmentation.
\newblock In {\em Proceedings of the IEEE conference on computer vision and
  pattern recognition}, pages 8759--8768, 2018.

\bibitem{logan2000mel}
Beth Logan.
\newblock Mel frequency cepstral coefficients for music modeling.
\newblock In {\em In International Symposium on Music Information Retrieval}.
  Citeseer, 2000.

\bibitem{loshchilov2017decoupled}
Ilya Loshchilov and Frank Hutter.
\newblock Decoupled weight decay regularization.
\newblock {\em arXiv preprint arXiv:1711.05101}, 2017.

\bibitem{lu2018listen}
Rui Lu, Zhiyao Duan, and Changshui Zhang.
\newblock Listen and look: Audio--visual matching assisted speech source
  separation.
\newblock {\em IEEE Signal Processing Letters}, 25(9):1315--1319, 2018.

\bibitem{milletari2016v}
Fausto Milletari, Nassir Navab, and Seyed-Ahmad Ahmadi.
\newblock V-net: Fully convolutional neural networks for volumetric medical
  image segmentation.
\newblock In {\em 2016 fourth international conference on 3D vision (3DV)},
  pages 565--571. IEEE, 2016.

\bibitem{mittal2020emotions}
Trisha Mittal, Uttaran Bhattacharya, Rohan Chandra, Aniket Bera, and Dinesh
  Manocha.
\newblock Emotions don't lie: An audio-visual deepfake detection method using
  affective cues.
\newblock In {\em Proceedings of the 28th ACM international conference on
  multimedia}, pages 2823--2832, 2020.

\bibitem{morrone2019face}
Giovanni Morrone, Sonia Bergamaschi, Luca Pasa, Luciano Fadiga, Vadim
  Tikhanoff, and Leonardo Badino.
\newblock Face landmark-based speaker-independent audio-visual speech
  enhancement in multi-talker environments.
\newblock In {\em ICASSP 2019-2019 IEEE International Conference on Acoustics,
  Speech and Signal Processing (ICASSP)}, pages 6900--6904. IEEE, 2019.

\bibitem{oh2019video}
Seoung~Wug Oh, Joon-Young Lee, Ning Xu, and Seon~Joo Kim.
\newblock Video object segmentation using space-time memory networks.
\newblock In {\em Proceedings of the IEEE/CVF International Conference on
  Computer Vision}, pages 9226--9235, 2019.

\bibitem{pu2017audio}
Jie Pu, Yannis Panagakis, Stavros Petridis, and Maja Pantic.
\newblock Audio-visual object localization and separation using low-rank and
  sparsity.
\newblock In {\em 2017 IEEE International Conference on Acoustics, Speech and
  Signal Processing (ICASSP)}, pages 2901--2905. IEEE, 2017.

\bibitem{qi2021occluded}
Jiyang Qi, Yan Gao, Yao Hu, Xinggang Wang, Xiaoyu Liu, Xiang Bai, Serge
  Belongie, Alan Yuille, Philip~HS Torr, and Song Bai.
\newblock Occluded video instance segmentation.
\newblock {\em arXiv preprint arXiv:2102.01558}, 2021.

\bibitem{schmidt1986multiple}
Ralph Schmidt.
\newblock Multiple emitter location and signal parameter estimation.
\newblock {\em IEEE transactions on antennas and propagation}, 34(3):276--280,
  1986.

\bibitem{senocak2018learning}
Arda Senocak, Tae-Hyun Oh, Junsik Kim, Ming-Hsuan Yang, and In~So Kweon.
\newblock Learning to localize sound source in visual scenes.
\newblock In {\em Proceedings of the IEEE Conference on Computer Vision and
  Pattern Recognition}, pages 4358--4366, 2018.

\bibitem{urvos}
Seonguk Seo, Joon-Young Lee, and Bohyung Han.
\newblock Urvos: Unified referring video object segmentation network with a
  large-scale benchmark.
\newblock In {\em European Conference on Computer Vision}, pages 208--223.
  Springer, 2020.

\bibitem{simonyan2014very}
Karen Simonyan and Andrew Zisserman.
\newblock Very deep convolutional networks for large-scale image recognition.
\newblock {\em arXiv preprint arXiv:1409.1556}, 2014.

\bibitem{tian2018audio}
Yapeng Tian, Jing Shi, Bochen Li, Zhiyao Duan, and Chenliang Xu.
\newblock Audio-visual event localization in unconstrained videos.
\newblock In {\em Proceedings of the European Conference on Computer Vision
  (ECCV)}, pages 247--263, 2018.

\bibitem{tian2019fcos}
Zhi Tian, Chunhua Shen, Hao Chen, and Tong He.
\newblock Fcos: Fully convolutional one-stage object detection.
\newblock In {\em Proceedings of the IEEE/CVF international conference on
  computer vision}, pages 9627--9636, 2019.

\bibitem{vaswani2017attention}
Ashish Vaswani, Noam Shazeer, Niki Parmar, Jakob Uszkoreit, Llion Jones,
  Aidan~N Gomez, {\L}ukasz Kaiser, and Illia Polosukhin.
\newblock Attention is all you need.
\newblock In {\em Advances in neural information processing systems}, pages
  5998--6008, 2017.

\bibitem{wang2020max}
Huiyu Wang, Yukun Zhu, Hartwig Adam, Alan Yuille, and Liang-Chieh Chen.
\newblock Max-deeplab: End-to-end panoptic segmentation with mask transformers.
\newblock {\em arXiv preprint arXiv:2012.00759}, 2020.

\bibitem{wang2020axial}
Huiyu Wang, Yukun Zhu, Bradley Green, Hartwig Adam, Alan Yuille, and
  Liang-Chieh Chen.
\newblock Axial-deeplab: Stand-alone axial-attention for panoptic segmentation.
\newblock In {\em European Conference on Computer Vision}, pages 108--126.
  Springer, 2020.

\bibitem{wang2020solo}
Xinlong Wang, Tao Kong, Chunhua Shen, Yuning Jiang, and Lei Li.
\newblock Solo: Segmenting objects by locations.
\newblock In {\em European Conference on Computer Vision}, pages 649--665.
  Springer, 2020.

\bibitem{wang2020solov2}
Xinlong Wang, Rufeng Zhang, Tao Kong, Lei Li, and Chunhua Shen.
\newblock Solov2: Dynamic, faster and stronger.
\newblock {\em arXiv preprint arXiv:2003.10152}, 2020.

\bibitem{wang2021end}
Yuqing Wang, Zhaoliang Xu, Xinlong Wang, Chunhua Shen, Baoshan Cheng, Hao Shen,
  and Huaxia Xia.
\newblock End-to-end video instance segmentation with transformers.
\newblock In {\em Proceedings of the IEEE/CVF Conference on Computer Vision and
  Pattern Recognition}, pages 8741--8750, 2021.

\bibitem{wang2019ranet}
Ziqin Wang, Jun Xu, Li Liu, Fan Zhu, and Ling Shao.
\newblock Ranet: Ranking attention network for fast video object segmentation.
\newblock In {\em Proceedings of the IEEE/CVF International Conference on
  Computer Vision}, pages 3978--3987, 2019.

\bibitem{referformer}
Jiannan Wu, Yi Jiang, Peize Sun, Zehuan Yuan, and Ping Luo.
\newblock Language as queries for referring video object segmentation.
\newblock {\em arXiv preprint arXiv:2201.00487}, 2022.

\bibitem{wu2021seqformer}
Junfeng Wu, Yi Jiang, Wenqing Zhang, Xiang Bai, and Song Bai.
\newblock Seqformer: a frustratingly simple model for video instance
  segmentation.
\newblock {\em arXiv preprint arXiv:2112.08275}, 2021.

\bibitem{xie2020polarmask}
Enze Xie, Peize Sun, Xiaoge Song, Wenhai Wang, Xuebo Liu, Ding Liang, Chunhua
  Shen, and Ping Luo.
\newblock Polarmask: Single shot instance segmentation with polar
  representation.
\newblock In {\em Proceedings of the IEEE/CVF conference on computer vision and
  pattern recognition}, pages 12193--12202, 2020.

\bibitem{xu2020cross}
Haoming Xu, Runhao Zeng, Qingyao Wu, Mingkui Tan, and Chuang Gan.
\newblock Cross-modal relation-aware networks for audio-visual event
  localization.
\newblock In {\em Proceedings of the 28th ACM International Conference on
  Multimedia}, pages 3893--3901, 2020.

\bibitem{yang2019video}
Linjie Yang, Yuchen Fan, and Ning Xu.
\newblock Video instance segmentation.
\newblock In {\em Proceedings of the IEEE/CVF International Conference on
  Computer Vision}, pages 5188--5197, 2019.

\bibitem{yang2018efficient}
Linjie Yang, Yanran Wang, Xuehan Xiong, Jianchao Yang, and Aggelos~K
  Katsaggelos.
\newblock Efficient video object segmentation via network modulation.
\newblock In {\em Proceedings of the IEEE Conference on Computer Vision and
  Pattern Recognition}, pages 6499--6507, 2018.

\bibitem{yang2021crossover}
Shusheng Yang, Yuxin Fang, Xinggang Wang, Yu Li, Chen Fang, Ying Shan, Bin
  Feng, and Wenyu Liu.
\newblock Crossover learning for fast online video instance segmentation.
\newblock {\em arXiv preprint arXiv:2104.05970}, 2021.

\bibitem{yang2019deeperlab}
Tien-Ju Yang, Maxwell~D Collins, Yukun Zhu, Jyh-Jing Hwang, Ting Liu, Xiao
  Zhang, Vivienne Sze, George Papandreou, and Liang-Chieh Chen.
\newblock Deeperlab: Single-shot image parser.
\newblock {\em arXiv preprint arXiv:1902.05093}, 2019.

\bibitem{zhao2018sound}
Hang Zhao, Chuang Gan, Andrew Rouditchenko, Carl Vondrick, Josh McDermott, and
  Antonio Torralba.
\newblock The sound of pixels.
\newblock In {\em Proceedings of the European conference on computer vision
  (ECCV)}, pages 570--586, 2018.

\bibitem{zhu2020deformable}
Xizhou Zhu, Weijie Su, Lewei Lu, Bin Li, Xiaogang Wang, and Jifeng Dai.
\newblock Deformable detr: Deformable transformers for end-to-end object
  detection.
\newblock {\em arXiv preprint arXiv:2010.04159}, 2020.

\end{thebibliography}
}
\clearpage

\title{Online Video Instance Segmentation via Robust Context Fusion}

\maketitle

\section{Lipschitz Continuity of Our Network}
We first define the notion of Lipschitz continuity, and proceed to the Lipschitz continuity of fully-connected, convolutional and self-attention layers. We refer to the proof in \cite{kim2021lipschitz} with bounded inputs.

\paragraph{High-level Proof}
Continuous functions are Lipschitz continuous given bounded inputs w.r.t. $p$-norm where $p\in[1, \infty]$.

\begin{definition}
\label{def:lip}
Given two metric spaces $(\mathcal{X}, d_\mathcal{X})$ and $(\mathcal{Y}, d_\mathcal{Y})$, a function $f:\mathcal{X}\rightarrow\mathcal{Y}$ is called \rm{Lipschitz continuous} (or \rm{K-Lipschitz}) if there exists a constant $K\leq0$ such that
\begin{equation}
    d_{\mathcal{Y}}(f(x), f(x^{\prime})) \leq Kd_{\mathcal{X}}(x, x^\prime)
\end{equation}
The smallest such $K$ is the Lipschitz constant of $f$, denoted $\mathrm{Lip}(f)$.
\end{definition}
In the following proof, the $d_{\mathcal{X}}$ and $d_{\mathcal{Y}}$ are induced by a $p$-norm $\|x\|_p:=(\sum_i|x_i|^p)^{1/p}$. Following \cite{kim2021lipschitz}, we emphasise the dependence of the Lipschitz constant on the choice of $p$-norm by denoting it as $\mathrm{Lip}_p(f)$. In this case, it follows directly from Definition~\ref{def:lip} that the Lipschitz constant is given by

\begin{equation}
    \mathrm{Lip}_p(f)=\sup_{x\neq x^\prime\in\mathbb{R}^n}\frac{\|f(x)-f(x^\prime)\|_p}{\|x-x^\prime\|_p}
\end{equation}

\subsection{Fully-connected/convolutional Layers}
We describe how Lipschitz constant of fully-connected networks (FC) and convolutional neural networks (CNN) are bounded, using the fact that both are compositions of linear maps and pointwise non-linearities. 
\begin{theorem}
(\cite{federer2014geometric}) Let $f: \mathbb{R}^n\rightarrow\mathbb{R}^m$ be differentiable and Lipschitz continuous under a choice of p-nom $\|\cdot\|_p$. Let $J_f(x)$ denote its total derivative (Jacobian at x. Then $\mathrm{Lip}_p(f)=\sup_{x\in\mathbb{R}^n}\|J_f(x)\|_p$) where $\|J_f(x)\|_p$ is the induced operator norm on $J_f(x)$.
\end{theorem}

Therefore if $f$ is a linear map represented by a matrix $W$ when
$$
    \mathrm{Lip}_p(f)=\|W\|_p:=\sup_{\|x\|_p=1}\|Wx\|_p
$$
$$
=\left\{
\begin{array}{lcr}
&\sigma_{max}(W) &if\, p=2 \\
&\max_i\sum_j|W_{ij}| &if\, p=\infty
\end{array}
\right.
$$
where $\|W\|_p$ is the operator norm on matrices induced by the vector $p$-norm, and $\sigma_{max}(W)$ is the largest singular value of $W$. 

We can show the Lipschitz continuity of FC and CNN by applying the following lemma.
\begin{lemma}
\label{lem:composition}
(\cite{federer2014geometric}) Let $g$, $h$ be two composable Lipschitz functions. Then $g\circ h$ is also Lipschitz with $\mathrm{Lip}(g\circ h)\leq \mathrm{Lip}(g)\mathrm{Lip}(h)$.
\end{lemma}

\begin{corollary}
For a fully-connected network or a convolutional neural network $f=W_k\circ \rho_{K-1}\circ W_{K-1}\circ\cdots\circ\rho_1 \circ W_1$, we have $\mathrm{Lip}_p(f)\leq\prod_k\|W_k\|_p$ under a choice of $p$-norm with 1-Lipschitz non-linearities $\rho_k$.
\end{corollary}

\subsection{Attention Layers}
The self-attention layer is not Lipschitz continuous \cite{kim2021lipschitz}. However, when we consider the bounded inputs, the Lipschitz constant of attention layers $\mathrm{Lip}_p(f)$ is bounded \cite{kim2021lipschitz}. Since the input images are normalized and the network is trained with adamw \cite{loshchilov2017decoupled} which penalizes the large weights, the inputs to attention layer is bounded. Therefore, we can assume the attention layer in our network is bounded with normalized image inputs.

\subsection{Entire Transformer-based Network}
Our transformer-based network is a composition of CNN, FC, and attention layers. By leveraging Lemma~\ref{lem:composition} on the normalized image space $\mathcal{I}$, we have
$$
\mathrm{Lip}_p(\mathrm{Model})\leq\mathrm{Lip}_p(\mathrm{CNN})\cdot\mathrm{Lip}_p(\mathrm{FC})\cdot\mathrm{Lip}_p(\mathrm{Attn})
$$
where $\mathrm{Lip}_p(\mathrm{Model})$, $\mathrm{Lip}_p(\mathrm{CNN})$, $\mathrm{Lip}_p(\mathrm{FC})$ and $\mathrm{Lip}_p(\mathrm{Attn})$ are the Lipschitz constants of the entire model, all CNN layers, all FC layers and all attention layers respectively.

\section{Audio-visual Correspondence}

\begin{figure}[h]
\centering
\includegraphics[width=\linewidth]{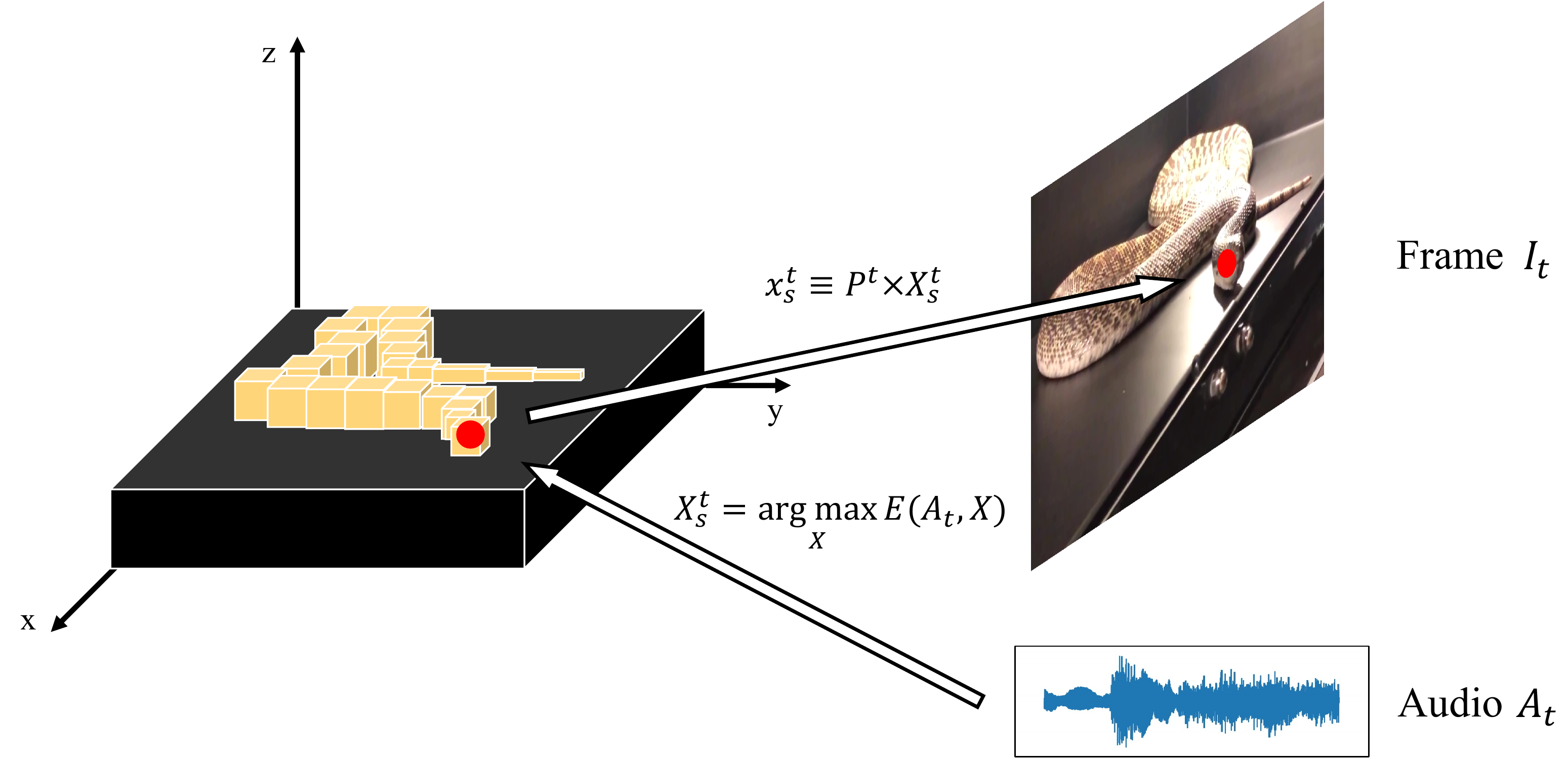}
\caption{\textbf{Audio-visual relationship.} }
\label{fig:av}
\end{figure}

Given the audio recordings $A^t$, we can decode the 3D location of sound source by signal processing methods such as MUSIC algorithm \cite{schmidt1986multiple} which searches for the location of the maximum of the spatial energy spectrum as
\begin{equation}
    X^t_s = \mathop{\arg \max}_{X} E(A^t, X)
\end{equation}
where $E(\cdot)$ is the spatial energy spectrum and $X^t_s\in\mathbb{R}^{3}$ is the 3D coordinate of sound source at time $t$. Therefore, the sound source can be easily corresponded to the pixel on a frame by applying camera matrix. Given that, we can represent the relationship between audio recordings $A^t$ and 2D location of sound source $x_s$ in frame $I_t$ by 
\begin{equation}
    x^t_s \equiv P^t\times \mathop{\arg \max}_{X} E(A^t, P)
\end{equation}
where $P^t$ is the camera matrix at time t and $x_t$ is the homogeneous coordinate of sound source in frame $I_t$. In practice, the camera matrix $P^t$ is unknown but can be estimated by a adjacent input frames $P^t=\mathcal{H}(I_t, I_{t-1})$. Consequently, we can link the sound source location $p^t_s$ in frames $I_t$ from audio $A_t$. 

\begin{figure*}[htbp]
\centering
\includegraphics[width=\linewidth]{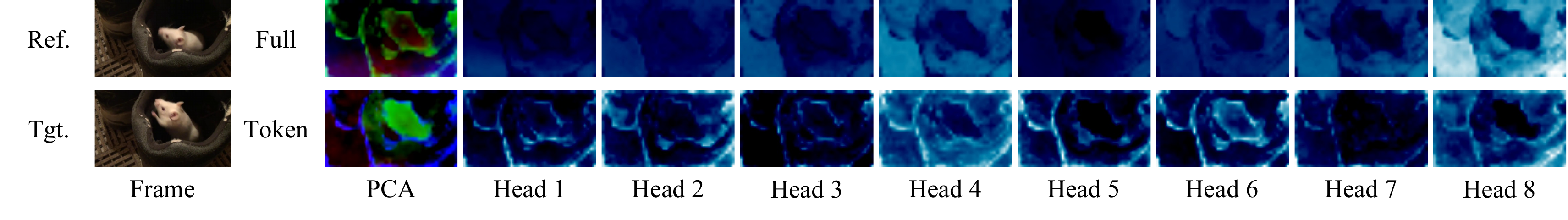}
\caption{\textbf{Visualization of reference-to-target attention map in last layer of transformer encoder.} We compare the attention map in the transformer encoder using full reference feature maps and compressed reference token. The PCA is conducted among the 8 heads of reference-to-target (r2t) attention map and keeps 3 main components for RGB channels. We only show the all 8 heads of the r2t attention. The attention map of token inputs is sharper on the instance region than that of full inputs which can also be verified by the PCA colormap. 
}
\label{fig:vis_mh_attn}
\end{figure*}

\section{Detailed Audio Preprocessing}
We first resample $A_i$ to 16 kHz mono then compute the spectrum using magnitudes of the Short-Time Fourier Transform (STFT) with a window size of 25 ms, a window hop of 10 ms, and a periodic Hann window. We pad the audio sequence to ensure the output has the same length as the input. The mel spectrogram is computed by mapping the spectrogram to 64 mel bins covering the range 125-7500 Hz. A stabilized log mel spectrogram is further computed by applying log(mel-spectrum + 0.01) where the offset is used to avoid taking a logarithm of zero.

\begin{table}[h]
\centering
\begin{minipage}[t]{0.4\textwidth}
\makeatletter\def\@captype{table}
\centering
\scalebox{0.7}{
    \begin{tabular}{ccccccc}
    \hline
    \hline
    Method & Backbone &AP & AP\@50 & AP\@75 & AR\@1 & AR\@10 \\
    \hline
    CrossVIS & ResNet-50 & 42.2 & 59.1 & 47.4 & 42.1 & 49.9 \\
    CrossVIS & ResNet-101 & 44.7 & 60.4 & 50.5 & 42.0 & 49.7 \\
    \textbf{Ours} & ResNet-50 & 46.6 & 63.3 & 51.3 & 44.2 & 51.5 \\
    \textbf{Ours} & ResNet-101 & 48.6 & 63.9 & 52.5 & 44.6 & 52.9 \\
    \hline
    \hline
    \end{tabular}}
    \caption{Comparison to state-of-the-art video instance segmentation on \textbf{AVIS} val set.}
    \label{tab:avis main}
\end{minipage}
\end{table}

\section{Additional Experiments on AVIS Dataset}
We report the performance of using ResNet-50 and ResNet-101 backbone on AVIS dataset. As shown in Table~\ref{tab:avis main}, we also compare the CrossVIS baseline which only leverages the video data. Our method outperforms CrossVIS baseline by 4.4 and 3.9 mAP with ResNet-50 and ResNet-101 backbone respectively.

\begin{table}
\centering
\begin{minipage}[h]{\linewidth}
\makeatletter\def\@captype{table}
\centering
\scalebox{0.75}{
    \begin{tabular}{c|cc|cc|c}
    \hline
    \hline
    Ref. number & AP & AP50 & AP & AP50 & p-value \\
    \hline
    & \multicolumn{2}{c|}{w/o Audio-token} & \multicolumn{2}{c|}{w Audio-token}\\
    \hline
    1 & 42.9 & 60.2 & 44.8 (+1.9) & 60.3 & 0.30 (0.24) \\
    2 & 44.3 & 60.9 & 46.0 (+1.7) & 63.8 & 0.26 (0.16) \\
    3 & 45.6 & 61.9 & 46.6 (+1.0) & 63.3 & 0.65 (0.42) \\
    4 & 45.6 & 62.2 & 45.7 (+0.1) & 62.7 & -\\
    5 & 45.2 & 62.9 & 45.0 (-0.2) & 62.6 & -\\
    \hline
    \hline
    \end{tabular}}
    \caption{Impact of integrating audio-token with different reference frames on \textbf{AVIS} val set. The p-value is computed by t-test among samples having an $\mathrm{IoU>0.5}$ ($\mathrm{IoU>0.75}$ in the bracket).}
    \label{tab:avis}
\end{minipage}
\end{table}

We also report the p-value of t-test on AVIS results with different thresholds. As shown in Table~\ref{tab:avis}, the p-value under IoU$>$0.75 threshold is smaller than that under IoU$>$0.5 threshold, which means the audio inputs have more effect on the high-quality results. 

\section{Difference with MaskFormer \cite{cheng2021per}}
MaskFormer \cite{cheng2021per} is an image-level panoptic segmentation method that treats prediction belonging to both stuff and things classes as a mask with corresponding semantic categories. The pixel and transformer decoder used in our model is similar to the structure adopted in MaskFormer. Here we formally analyze the difference between our method and MaskFormer.

Our method is a video-level instance segmentation method which equips with a context fusion module (CFM) to fuse the contextual information. The proposed CFM module is leveraged to compress the historical features from adjacent frames with low redundancy and construct the spatial correlation between target and contextual features using attention mechanism. The importance of the final segmentation is considered when fusing the context features. By utilizing the flexibility of CFM, we also fuse the audio modality to facilitate the instance segmentation task. The transformer decoder of our model takes the multi-modal features as inputs, which is different from MaskFormer. The decoded instance embedding is further to be leveraged to track the instance identities given the Lipschitz continuity of the network.



\section{Visualization of Attention Maps in CFM}
We demonstrate the attention map of all eight heads of the reference-to-target attention. The attention map of the compressed feature is sharper than the attention map using the full features. The attention map is selected from the same position in the transformer encoder for fairly comparison.

\section{Limited Gain of Involving Audio Modality}
Our method introduces audio modality in an online fashion with synchronized the audio and video data. Since we considering streamlining video frames (as previous online VIS methods), we also introduce the corresponding audio frames in the same way. However, for audio data analysis, long-term historical information is required. Although we have reference audio frames, they are limited to a small number since we can only consider a moderate number of reference video frames with speed concerns. In this way, we believe that the audio modality can be further leveraged in asynchronized audio-visual inputs or clip-level inputs. We will investigate those settings in the future.



\end{document}